%% file: neurips_2026.tex
\documentclass{article}

\PassOptionsToPackage{numbers,sort&compress}{natbib}
 \usepackage[preprint]{neurips_2026}

\usepackage[utf8]{inputenc} 
\usepackage[T1]{fontenc}    
\usepackage{hyperref}       
\usepackage{url}            
\usepackage{booktabs}       
\usepackage{amsfonts}       
\usepackage{nicefrac}       
\usepackage{microtype}      
\usepackage{xcolor}         

\input{macros.tex}

\title{Diversify Diffusion with Temperature Sampling and Variance-Corrective Time Shifting}
\author{%
  Peizhuo Li \\
  ETH Zurich \\
  \And
  Emre Aksan \\
  Google \\
  \And
  Alexandru-Eugen Ichim \\
  Google \\
  \AND
  Thabo Beeler \\
  Google \\
  \And
  Olga Sorkine-Hornung \\
  ETH Zurich \\
}

\begin{document}

\maketitle

\begin{figure*}[h]
\input{figures/0-teaser.tex}
\end{figure*}

\begin{abstract}
\input{0-abstract.tex}
\end{abstract}

\input{1-intro.tex}

\input{2-relatedworks.tex}
\input{4-method.tex}
\input{5-experiments.tex}
\input{6-conclusion.tex}

\bibliographystyle{plainnat}
\bibliography{bibs.bib}


\appendix

\input{7-appendix.tex}


\end{document}

%% file: macros.tex
\usepackage{hyperref}
\usepackage{amsmath}
\usepackage{graphicx}
\usepackage{comment}
\usepackage{multirow}
\usepackage{cleveref}
\usepackage{dsfont}
\usepackage{mathtools}

\crefname{equation}{Eq.}{Eqs.}
\Crefname{equation}{Eq.}{Eqs.}

\crefname{section}{Sec.}{Secs.}
\Crefname{section}{Sec.}{Secs.}

\crefname{figure}{Fig.}{Figs.}
\Crefname{figure}{Fig.}{Figs.}

\crefname{table}{Tab.}{Tabs.}
\Crefname{table}{Tab.}{Tabs.}

\Crefname{appendix}{App.}{Apps.}
\crefname{appendix}{App.}{Apps.}

\usepackage{array}
\usepackage{wrapfig}

\usepackage{csvsimple}
\usepackage{adjustbox}
\usepackage[percent]{overpic}

\DeclareMathOperator*{\argmin}{arg\,min}

\usepackage{color}
\usepackage{soul}
\usepackage{breqn}
\usepackage{booktabs}
\usepackage{tabularx}
\usepackage{rotating}

\definecolor{purple}{rgb}{0.99,0.2,0.72}
\definecolor{blue}{rgb}{0, 0.2, 0.8}
\definecolor{orange}{rgb}{0.6, 0.6, 0}
\definecolor{red}{rgb}{0.8, 0.2, 0.2}
\definecolor{magenta}{rgb}{0.5, 0.0, 1.0}
\definecolor{black}{rgb}{0.0, 0.0, 0.0}
\definecolor{cyan}{rgb}{0, 0.65, 0.65}
\definecolor{olive}{rgb}{0.2, 0.6, 0.5}
\usepackage{xcolor}

\newcommand{\TempParam}{\gamma} 
\newcommand{\TempDistSym}{p} 
\newcommand{\TempDistSup}{^{(\TempParam)}} 
\newcommand{\TempDist}{\TempDistSym\TempDistSup} 
\newcommand{\TempDistT}[1]{\TempDistSym_{#1}\TempDistSup} 
\newcommand{\TempDistTX}[2]{\TempDistT{#1}\!\left(#2\right)} 
\newcommand{\TempNorm}{Z_{\TempParam}} 

\newif\ifdraft
\drafttrue

\ifdraft
    \newcommand{\aic}[1]{{\color{olive}[\textbf{Alex:} \textit{#1}]}}
    \newcommand{\plc}[1]{{\color{cyan}[\textbf{Peizhuo:} \textit{#1}]}}
    \newcommand{\OSHc}[1]{{\color{purple}[\textbf{Olga:} \textit{#1}]}}

    \renewcommand{\aic}[1]{{\color{magenta}#1}}

\else
    \newcommand{\aic}[1]{}
    \newcommand{\plc}[1]{}
    \newcommand{\OSHc}[1]{}
    \renewcommand{\aic}[1]{{\color{black}#1}}

\fi

\usepackage{pgfplots}
\pgfplotsset{compat=newest}
\usepgfplotslibrary{groupplots}
\usepgfplotslibrary{dateplot}

\usepackage[bottom]{footmisc}
\usepackage{units}

%% file: figures/0-teaser.tex
    \centering
    \includegraphics[width=\linewidth]{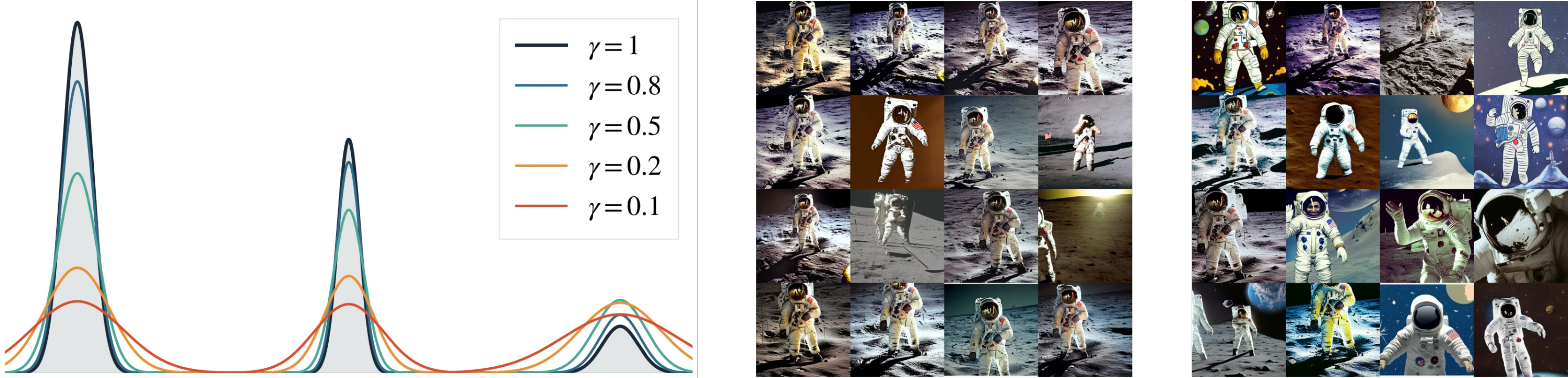}
    \caption{Left: The effect of increasing temperature (\emph{i.e.} lowering $\gamma$) on a 1D Gaussian mixture. Middle: Sampling results with prompt ``\emph{an astronaut on the moon}'' using Stable Diffusion. Right: The effect of our method on the same prompt and model.}
    \label{fig:teaser}

%% file: 0-abstract.tex
Diffusion models faithfully reproduce their training distribution, but also inherit its imbalances and leave rare or under-represented modes hard to reach.
A natural inference-time remedy is to sample from the high-temperature target $\TempDist_0(x) \propto p_0(x)^{\TempParam}$ for $0 < \TempParam < 1$, which flattens dominant modes and lifts rare ones. However, naive score scaling while correctly reweighting modes also inflates the per-mode variance, breaking the reverse diffusion process and degrading sample quality.
We introduce \emph{variance-corrective time shifting}, a training-free fix that queries the network at a shifted timestep and scales the resulting score by $\TempParam$, canceling the variance inflation while preserving the mode reweighting. The correction turns simple temperature sampling into a practical diversity knob for pretrained diffusion and flow-matching backbones with no retraining, and we demonstrate consistent gains at minimal cost to sample quality and condition fidelity across DiT, Stable Diffusion and Motion Diffusion models.
We further show that the timing of the temperature intervention enables coarse-to-fine control: high-noise stages drive compositional diversity across modes, while low-noise stages drive local appearance variation under a fixed composition.

%% file: 1-intro.tex
\section{Introduction}

Diffusion models~\cite{ho2020denoising, sohl2015deep, song2020score} have emerged as the dominant paradigm for generative modeling across images, video and 3D content. Beyond their high quality output, another key strength of diffusion models is that they offer a flexible inference-time control mechanism~\cite{ho2022classifier}, which allows users to steer the sampling process without retraining the model.
Yet, their samples often reflect the biases of the data they are trained on: common modes are reproduced reliably, while rare or under-represented modes are unlikely to be generated.
Although this behavior aligns the goal for matching the training distribution, the lack of diversity could be problematic in creative settings, where users care about exploring multiple plausible outcomes rather than a single output.

A natural way to surface rare modes is to draw from the high-temperature target
\begin{equation}
\TempDist_0(x) \coloneqq \frac{p_0(x)^{\TempParam}}{\TempNorm},
\qquad
\TempNorm \coloneqq \int p_0(x)^{\TempParam} \, \mathrm{d}x,
\label{eq:temp-dist-def}
\end{equation}
which flattens dominant modes and boosts the relative probability of rare events for $0 < \TempParam < 1$, as shown in \Cref{fig:teaser} (left). Here, $p_0(x)$ is the original data distribution and $\TempParam$ is the \emph{inverse temperature} parameter that controls the strength of the effect, with $\TempParam=1$ recovering the original distribution and $\TempParam \to 0$ approaching a uniform distribution, \emph{i.e.} infinitely high temperature. 
In diffusion models, a neural network learns the score, the logarithmic likelihood gradient, and gradually denoises the random distribution $p_T(x)$ to the data distribution $p_0(x)$ guided by the score. The temperature sampling looks deceptively simple on score: the identity $\nabla_x \log \TempDist_0(x) = \TempParam \, \nabla_x \log p_0(x)$ holds strictly and conveniently cancels the unknown normalizer $\TempNorm$. One might therefore hope to obtain $\TempDist$ by multiplying the learned score by $\TempParam$ throughout the reverse diffusion process.

However, naive score scaling conflates two effects. It achieves the \emph{macroscopic} goal of reweighting modes, but it also induces an unwanted \emph{microscopic} variance expansion that introduces more noise than expected, as demonstrated in \Cref{fig:teaser} (left). The excess variance breaks the strict correspondence between the score of the noised tempered marginal $\nabla_x \log \TempDistT{t}(x)$ and the reverse diffusion process, as shown in \Cref{sec:variance-time-shifting}. It lowers sample quality and produces off-manifold and blurry samples.

To mitigate this issue, we introduce \emph{variance-corrective time shifting}, a training-free fix that decouples the two effects: rather than querying the network at the true timestep $t$, we query it at a shifted timestep $\tilde{t}$ whose smaller noise level cancels the variance inflation. Concretely, presenting the model with the noisy sample $x_t$ at a smaller noise level and scaling the resulting score by $\TempParam$ lands the effective conditional variance back on the original schedule while preserving the mode reweighting. This cancelation is a very accurate approximation under an assumption of well-separated modes on the data (\Cref{sec:variance-time-shifting}), which empirically performs well for  natural images and motion generation considered in our experiments.

We show that with this method, sampling diversity is improved on a range of models, including DiT~\cite{peebles2023scalable}, Stable Diffusion~\cite{rombach2022high, podell2023sdxl, esser2024scaling} and Motion Diffusion Model~\cite{tevet2022human}, at minimal cost to sample quality and condition fidelity. We also explore the effects of temperature scaling on the different stages of the diffusion process. Similarly to the natural coarse-to-fine reverse diffusion process, applying $\TempParam(t) < 1$ at early (high-noise) timesteps yields compositional diversity across modes, while applying it at later (low-noise) timesteps yields local appearance variation under a fixed composition.

In summary, our contributions are:
\begin{itemize}
    \item We enable simple temperature sampling as a training-free mechanism for enhancing diversity in pretrained diffusion models, where naive application fails to produce usable samples.
    \item We identify the \emph{variance expansion problem} of naive temperature sampling in diffusion models, which breaks the reverse diffusion process and degrades sample quality.
    \item We propose \emph{variance-corrective time shifting}, a zero-cost correction that recovers the desired tempered marginal by querying the network at a shifted timestep.
    \item We demonstrate the effectiveness of our method across various models and modalities, showing consistent gains in diversity at minimal cost to sample quality and condition fidelity.
\end{itemize}

%% file: 2-relatedworks.tex
\section{Related work}
Diffusion models~\cite{ho2020denoising, sohl2015deep, song2020score, song2020denoising} have quickly achieved superior performance compared with GANs~\cite{goodfellow2014gan} and became the gold standard for generative modeling. Beyond the high quality itself, another key strength of diffusion models is the flexible inference-time control mechanism~\cite{dhariwal2021diffusion, ho2022classifier, bansal2023universal, skreta2025feynman}, which allows users to steer the sampling process without retraining the model.

Pioneered by \citet{kirkpatrick1983optimization}, temperature scaling or sampling has been used in various deep learning field~\cite{hinton2015distilling, shih2023long, holtzman2019curious, nijkamp2023progen2}. However, the majority of the work in generative modeling focuses on \emph{lowering} the temperature to sharpen the distribution and produce higher-quality samples. 
One exception is TSR~\cite{xu2025temporal}, which applies temperature scaling to control the variance of the output distribution. However, this method suffers from the mentioned variance expansion problem, and cannot be used to produce high-quality and diverse samples when raising the temperature. In contrast, our method is designed to mitigate this issue and enable effective high temperature sampling for diversity enhancement.

Several recent methods enhance diversity in pretrained diffusion models without retraining, each approaching the problem from a different angle.
\citet{sehwag2022generating} bias the sampler towards low-density regions of the data manifold, but they require a pre-trained classifier, whereas our approach is agnostic to data-distribution and requires no auxiliary annotation.
CADS~\cite{sadat2024cads} anneals the conditioning signal at high noise to broaden conditional generation, but it operates on the conditioning, rather than the data distribution itself, and therefore tends to produce samples that are less faithful to the condition.
Closer to our goal of sampling from $\TempDist_0$, Feynman-Kac correctors (FK)~\cite{skreta2025feynman} use sequential Monte Carlo particle filtering to correct the target distribution exactly, which is mathematically clean and distribution-agnostic but requires running a large batch of particles in parallel. Although it performs well in the annealed setting ($\TempParam > 1$), this method struggles with high temperatures ($\TempParam < 1$) in our experiments.
A parallel technique particle guidance~\cite{corso2024particle} adds a pairwise repulsive potential across particles in a sampled batch, inducing non-i.i.d.\ batch-level diversity. It operates \emph{across} particles, while our method operates \emph{within} each particle.

A separate line of work modifies the training-time formulation to better represent long-tailed data, but does not directly increase the probability of sampling from the tails.
Heavy-tailed diffusion~\cite{pandey2024heavy} replaces the Gaussian prior with a $t$-distribution, so that the model can capture rare events present in the training data. This improves how the long tail is \emph{represented} but does not change the relative sampling probabilities, so it does not on its own increase the chance of generating rare modes the way temperature sampling does.

%% file: 4-method.tex
\section{Method}

\subsection{Variance-corrective time shifting for temperature sampling}
\label{sec:variance-time-shifting}

For simplicity, we consider a variance-exploding (VE) diffusion with forward kernel $p(x_t\mid x_0) = \mathcal{N}(x_t; x_0, \sigma_t^2 I)$, where $\sigma_t$ is a noise schedule that increases with time. 
We work with unnormalized densities below, since normalizing constants do not affect the score function or the sampling trajectory.
Let $p_0(x)$ denote the data distribution and $p_t(x)$  the noisy marginal at time $t$. The reverse-time sampling process is defined by the SDE~\cite{song2020score}:
\begin{equation}
\mathrm{d}x = \left[ f(x,t) - g(t)^2 \nabla_x \log p_t(x) \right] \mathrm{d}t + g(t) \, \mathrm{d}w,
\end{equation}
where $f$ and $g$ are known functions determined by the forward process.
Now consider the tempered distribution $\TempDist_0$ obtained by raising $p_0$ to the power $\TempParam$ (and renormalizing), as defined in \Cref{eq:temp-dist-def}. Its corresponding unnormalized noisy marginal at time $t$ is
\begin{equation}
\TempDistTX{t}{x} \coloneqq \int p_0(x_0)^{\TempParam} \, \mathcal{N}(x; x_0, \sigma_t^2 I) \, \mathrm{d}x_0,
\label{eq:temp-noisy-marginal}
\end{equation}
while the original noisy marginal is
\begin{equation}
p_t(x) = \int p_0(x_0) \, \mathcal{N}(x; x_0, \sigma_t^2 I) \, \mathrm{d}x_0.
\label{eq:orig-noisy-marginal}
\end{equation}
One might instead apply tempering \emph{after} noising, by taking the (unnormalized) power
\begin{equation}
    p_t(x)^{\TempParam} = \left(\int p_0(x_0) \, \mathcal{N}(x; x_0, \sigma_t^2 I) \, \mathrm{d}x_0\right)^{\TempParam},
    \label{eq:direct-tempering}
\end{equation}
which we call \emph{direct temperature scaling}. In this case the score is obtained by simple scaling:
\begin{equation}
    \nabla_x \log p_t(x)^{\TempParam} = \TempParam \, \nabla_x \log p_t(x).
\end{equation}
However, direct temperature scaling is generally \emph{not} equivalent to noising the tempered data distribution, by comparing \Cref{eq:temp-noisy-marginal} and \Cref{eq:direct-tempering}. Thus, the score of $\TempDistTX{t}{x}$ cannot be obtained by simply scaling the original score:
\begin{equation}
    \nabla_x \log \TempDistTX{t}{x} \neq \TempParam \, \nabla_x \log p_t(x).
\end{equation}
Intuitively, $\TempDistTX{t}{x}$ reweights $p_0$ \emph{before} Gaussian convolution, whereas $p_t(x)^{\TempParam}$ applies the reweighting \emph{after} noising.
To further study this mismatch, we look at a simple example where $p_0$ is a multimodal distribution with well-separated modes, and we consider times $t$ for which $\sigma_t$ is small enough that the modes remain separated. In this regime,
\begin{equation}
    p_t(x) \approx \sum_{k} \pi_k \, \mathcal{N}(x; \mu_k, \sigma_t^2 I),
\end{equation}
where $\pi_k$ is the probability mass of the $k$-th mode under $p_0$ and $\mu_k$ is the corresponding mean.
Under this approximation, we can write down a closed-form expression for the unnormalized $\TempDistT{t}$:
\begin{equation}
    \TempDistT{t}(x) = \sum_k \pi_k^{\TempParam} \, \mathcal{N}(x; \mu_k, \sigma_t^2 I).
\end{equation}
In contrast, direct tempering of $p_t$ gives
\begin{equation}
    p_t(x)^{\TempParam} = \left(\sum_k \pi_k \, \mathcal{N}(x; \mu_k, \sigma_t^2 I)\right)^{\TempParam}.
\end{equation}
With the well-separated assumption, cross terms are negligible (see \Cref{app:cross-terms}), so
\begin{equation}
    p_t(x)^{\TempParam} \approx \sum_k \pi_k^{\TempParam} \, \mathcal{N}(x; \mu_k, \sigma_t^2 I)^{\TempParam} \propto \sum_k \pi_k^{\TempParam} \, \mathcal{N}(x; \mu_k, \sigma_t^2 I/\TempParam).
    \label{eq:cross-term-negligible}
\end{equation}
This motivates a simple time shift: choose $\tilde{t}$ such that $\sigma_{\tilde{t}}^2 = \TempParam\,\sigma_t^2$. Then
\begin{equation}
    p_{\tilde{t}}(x)^{\TempParam} \approx \sum_k \pi_k^{\TempParam} \, \mathcal{N}(x; \mu_k, \sigma_t^2 I).
\end{equation}
Assuming we have access to a pre-trained neural network $\epsilon_\theta(x,t)$ that predicts the perturbation noise ($\epsilon$-prediction), Tweedie's formula~\cite{vincent2011connection} yields a score estimator $\nabla_x \log p_t(x) = -\epsilon_\theta(x,t)/\sigma^2_t$. We can then approximate the score of $\TempDistT{t}$ by querying the network at the shifted time $\tilde{t}$ and applying temperature scaling:
\begin{equation}
    \nabla_x \log \TempDistT{t}(x) \approx \nabla_x \log\left(p_{\tilde{t}}(x)^{\TempParam}\right)
    = \TempParam \, \left(-\frac{\epsilon_\theta(x,\tilde{t})}{\sigma^2_{\tilde{t}}}\right),
    \label{eq:time-shifted-score}
\end{equation}
which we refer to as \emph{variance-corrective time shifting}. It effectively compensates the variance expansion while achieving the desired mode reweighting.

Although the derivation above is stated for variance-exploding diffusion, the same construction extends to other diffusion processes and samplers. We instantiate it for variance-preserving diffusion in \Cref{sec:appendix-vp} and for rectified-flow models in \Cref{app:flow-matching}.

\subsection{Toy example}

We validate our mode reweighting on a simple 2D Gaussian mixture,
\begin{equation}
    p_0(x) = \sum_{k=1}^3 \pi_k\,\mathcal{N}(x;\mu_k,\sigma_{\text{data}}^2 I),
\end{equation}
where the three centers $\{\mu_k\}_{k=1}^3$ are sampled once at random in $[-1.5,1.5]^2$, the mixture weights are $\pi=(0.8, 0.1, 0.1)$ and $\sigma_{\text{data}}$ is $0.01$.
We train a small $\epsilon$-prediction model on samples from $p_0$ and generate samples using our temperature parameter $\gamma$.

\input{figures/2-tab-toy-example.tex}

To quantify mode reweighting, for each $\gamma$ we draw $N=10{,}000$ samples with DDPM~\cite{ho2020denoising}, assign each sample to its nearest center, and report empirical reach probabilities $\hat p_k(\gamma)$ given by the resulting frequencies.
For reference, the ideal tempered mode weights are $\pi_k^{(\gamma)} \coloneqq \pi_k^{\gamma}/\sum_j\pi_j^{\gamma}$.
Table~\ref{tab:toy-reach-probs} summarizes the comparison between $\hat p(\gamma)$ and $\pi^{(\gamma)}$. It can be seen that with small $\gamma$, the empirical reach probabilities get closer to evenly distributed, while with large $\gamma$, the dominant modes are enhanced.

Although idealized, this toy setup captures a structural property shared by many real-world generative tasks. Natural data such as images or human motion can often be modeled as an unbalanced mixture of well-separated modes, where dominant categories or behaviors crowd out rarer ones. Under this approximation, the mode-reweighting effect demonstrated above predicts that lowering $\TempParam$ at sampling time will redistribute mass toward underrepresented modes, which is the mechanism we exploit to enhance diversity in the experiments that follow.

\subsection{Different stages of the diffusion process}

The reverse diffusion process is inherently coarse-to-fine: early denoising steps, operating at high noise levels, determine the global composition and mode identity of the generated sample, while later steps progressively refine structural details and texture. This suggests that the effect of tempering depends not just on its strength but on \emph{when} along the trajectory it is applied. To explore this empirically, we relax the constant-$\TempParam$ setting used so far and replace $\TempParam$ with a sampling-time schedule $\TempParam(t)$, applied step-by-step using the same time-shifting and score-scaling rule. We treat $\TempParam(t)$ as a practical knob that lets us control which portion of the denoising trajectory is reweighted.

\input{figures/1-multi-stage.tex}

To study how each stage contributes, we partition the $T$ sampling steps into three equal segments: Early ($T - \frac{2}{3}T$), Middle ($\frac{2}{3}T - \frac{1}{3}T$) and Late (steps $\frac{1}{3}T - 0$), where step $T$ is the noisiest. We choose $\TempParam(t) < 1$ inside one segment at a time while keeping $\TempParam(t) = 1$ elsewhere. 

We find the effect of temperature scaling to be qualitatively distinct across the three stages, as shown in \Cref{fig:multi-stage}. The \emph{Early} segment is where noise is high and the score encodes global structure. Applying $\TempParam(t) < 1$ here steers sampling particles towards different attractors, increasing overall compositional diversity and enabling the sampling of different modes. Applying it during the \emph{Middle} segment preserves the global composition and mode identity but introduces variation in mid-level structural details. Applying it during the \emph{Late} segment, when the score refines local appearance, yields samples with identical overall structure but differing  in fine-grained details.

%% file: figures/2-tab-toy-example.tex
\begin{table}[t]
    \caption{Toy example mode reweighting. For each $\gamma$ we compare the target tempered mode weights $\pi^{(\gamma)}$ to empirical reach probabilities $\hat p(\gamma)$ and report $\mathrm{KL}(\hat p(\gamma)\,\|\,\pi^{(\gamma)})$.}
  \label{tab:toy-reach-probs}
  \centering
  \small
  \setlength{\tabcolsep}{4pt}
  \scalebox{0.85}{
  \begin{tabular}{c cc cc cc c}
    \toprule
    & \multicolumn{2}{c}{mode 1} & \multicolumn{2}{c}{mode 2} & \multicolumn{2}{c}{mode 3} & \\
    \cmidrule(lr){2-3}\cmidrule(lr){4-5}\cmidrule(lr){6-7}
    $\gamma$ & $\pi_1^{(\gamma)}$ & $\hat p_1(\gamma)$ & $\pi_2^{(\gamma)}$ & $\hat p_2(\gamma)$ & $\pi_3^{(\gamma)}$ & $\hat p_3(\gamma)$ & $\mathrm{KL}$ \\
    \midrule
    0.1 & 0.381 & 0.464 & 0.309 & 0.227 & 0.309 & 0.309 & 0.021 \\
    0.5 & 0.586 & 0.640 & 0.207 & 0.167 & 0.207 & 0.192 & 0.007 \\
    1.0 & 0.800 & 0.794 & 0.100 & 0.091 & 0.100 & 0.115 & 0.001 \\
    5.0 & 1.000 & 0.999 & 0.000 & 0.001 & 0.000 & 0.001 & 0.003 \\
    \bottomrule
  \end{tabular}
  } 
\end{table}

%% file: figures/1-multi-stage.tex
\begin{figure}
    \centering
    \scalebox{0.98}{
    \def\panelY{-2.5}
    \begin{overpic}[width=\linewidth]{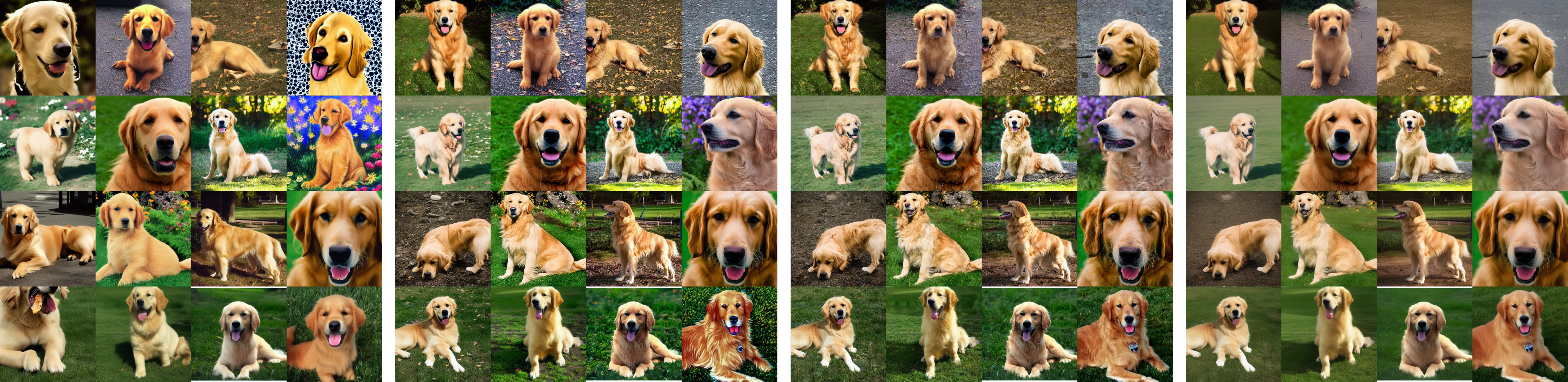}
        \put(12.5,\panelY){\makebox[0pt][c]{\small Early}}
        \put(37.5,\panelY){\makebox[0pt][c]{\small Middle}}
        \put(62.5,\panelY){\makebox[0pt][c]{\small Late}}
        \put(87.5,\panelY){\makebox[0pt][c]{\small No Tempering}}
    \end{overpic}
    }
    \vspace{0.1pt}
    \caption{The effect of applying temperature scaling at different denoising stages.
    }
    \label{fig:multi-stage}
\end{figure}

%% file: 5-experiments.tex
\section{Experiments}

\input{figures/4-comparison-all}

\subsection{Fidelity and diversity metrics}
\label{sec:metrics}

The CLIP Maximum Mean Discrepancy (CMMD)~\cite{jayasumana2024rethinking} is the modern standard for perceptual quality of generative models, as it uses richer CLIP embeddings and an unbiased kernel MMD estimator, making it more effective at detecting semantic distribution differences and substantially more computationally efficient than FID~\cite{heusel2017gans}. CMMD rewards a generator for matching the reference data distribution, whereas our goal is to deliberately deviate from the training distribution by sampling from a tempered version $\TempDist$ for diversity. As a result, CMMD penalizes our method for doing exactly what we aim for. We report CMMD for completeness and only as a partial reference for the quality of generated samples. For text-conditioned models, we report CMMD on the COCO~\cite{lin2014microsoft} validation set. For class-conditional models, we report CMMD on the ImageNet~\cite{deng2009imagenet} validation set. We use 2,000 samples for all CMMD calculations.

We report two complementary metrics that decouple condition fidelity from sample diversity, so the trade-off curve induced by varying $\TempParam$ can be read off directly.

\paragraph{Condition fidelity.} For text-conditioned image generation we use the VQA score~\cite{lin2024evaluating}, which scores prompt alignment by querying a vision--language model with the prompt rephrased as a yes/no question and reading out the probability of the affirmative answer. The VQA score has been shown to correlate well with human judgment of prompt--image alignment, in particular for compositional prompts where CLIP-based scores~\cite{radford2021learning} often saturate. For class-conditioned image generation we report the top-1 accuracy of a pretrained classifier on the generated samples.

\paragraph{Sample diversity.} Following CADS~\cite{sadat2024cads}, we quantify the diversity of a set of samples drawn from the same condition using the Vendi score~\cite{friedman2022vendi} (full definition in~\Cref{app:vendi}). Intuitively, the Vendi score reports the effective number of unique elements in a set: a score of $1$ means all samples are identical, while a score of $n$ (the size of the set) means the samples are maximally diverse under the chosen similarity kernel. Unlike recall- or coverage-based diversity measures, it does not require a reference distribution, which is appropriate here because the reference is exactly the distribution we are deviating from.

\subsection{Results and comparisons}
\label{sec:results}

\paragraph{Image generation.} In this seciton we compare our method with others, including CADS~\cite{sadat2024cads} and Feynman--Kac~\cite{skreta2025feynman}, on Stable Diffusion 1.5, Stable Diffusion XL, Stable Diffusion 3.5 and DiT. For the same model, all generations are conditioned on the same prompt, $\TempParam$ and seed where applicable. Since we are interested in mode diversity that is controled in the early stage of reverse diffusion, we apply the tempering only in the early stage for all methods throughout \Cref{sec:results}.

As shown in \Cref{fig:comparison-all}, our method consistently improves sample diversity across all models while preserving prompt fidelity. CADS degrades noticeably on SDXL and fails catastrophically on SD3.5, whereas Feynman--Kac injects excessive noise throughout generation and yields unusable samples. The gain on DiT is more modest, since the model is less expressive than the Stable Diffusion family and therefore offers less headroom for diversity gains without compromising fidelity.

\input{figures/8-comparison-table}

Quantitatively, our method attains the best CMMD and fidelity scores on all models and ranks first or second on Vendi, as reported in \Cref{tab:comparison}. CADS trades conditional fidelity for higher Vendi score and breaks down on SD3.5, whereas our method lifts diversity without sacrificing alignment. The qualitative comparison in \Cref{fig:comparison-all} further reveals greater stylistic variety in our outputs, an axis that the Vendi score may not fully reflect.

\paragraph{Motion generation.} Reliable metrics for diversity and quality in human motion generation~\cite{tevet2022human} are still being actively explored, so we opt to provide qualitative \emph{video} results in the supplementary material. We also include a static image in \Cref{fig:comparison-human}. The diversity gain is more apparent in video format, so we encourage readers to refer to the supplementary video for a full comparison. Our method produces more diverse motion patterns while maintaining the same overall pose and trajectory, whereas CADS produces motion that diverges from the text prompt, similar to the case of SD3.5 in image generation.

\subsection{Ablation study}

\paragraph{Choice of $\TempParam$ and stage.}

We sweep $\TempParam$ over a range of values for each model and compare three denoising stages (early, mid and late), and different conditioning text and class tiers (loose, medium, tight), depending on the diversity headroom. Full setup and conditioning sets are given in \Cref{app:prompts}.

We observe a consistent pattern as $\TempParam$ varies. At small $\TempParam$, samples are dominated by noise artifacts and appear unnatural, as illustrated in the upper-left corner of \Cref{fig:stage-gamma-grid}, which translates into low Vendi and fidelity scores. As $\TempParam$ grows, samples become more natural and the Vendi score rises, peaking before falling back toward the baseline as $\TempParam$ approaches $1$. Fidelity, in contrast, increases roughly monotonically with $\TempParam$. A second observation is that tempering at the early stage typically produces the largest diversity gains, with SDXL being the sole exception. Guided by \Cref{fig:gamma-sweep}, we select for each model the smallest $\TempParam$ that yields a substantial gain in Vendi without a meaningful drop in fidelity, giving $\TempParam = 0.20$ for DiT, $\TempParam = 0.88$ for SD1.5, $\TempParam = 0.94$ for SDXL and $\TempParam = 0.78$ for SD3.5. These are the values used in \Cref{sec:results}. A detailed per-tier breakdown is given in \Cref{fig:gamma-sweep-full}.
\input{figures/2b-gamma-sweep-avg}

\input{figures/3-stage-gamma-grid}

\paragraph{Variance-corrective time shifting.}

\input{figures/7-ddim-time-shift.tex}

Here, we make a comparison between our method without time shifting (i.e., direct tempering) and TSR~\cite{xu2025temporal}. TSR is effectively a direct tempering method with a smooth $\gamma(t)$ where $\gamma$ smoothly transits from 1 to the supplied value as the diffusion process progresses from noisy to clean. As shown in \Cref{fig:ddim-time-shift}, TSR introduces increasingly more local details and eventually becomes noisy as $\gamma$ decreases. Without time shifting, small $\TempParam$ values introduce excessive noise and little diversity due to the variance expansion problem. In contrast, our method with time shifting can maintain good sample quality while achieving the same level of diversity as direct tempering.





%% file: figures/4-comparison-all.tex
\begin{figure*}[t]
    \centering

    \makebox[0.25\linewidth][c]{\small CADS}%
    \makebox[0.25\linewidth][c]{\small FK}%
    \makebox[0.25\linewidth][c]{\small Ours}%
    \makebox[0.25\linewidth][c]{\small Baseline ($\gamma = 1$)}

    \vspace{2pt}

    \includegraphics[width=\linewidth]{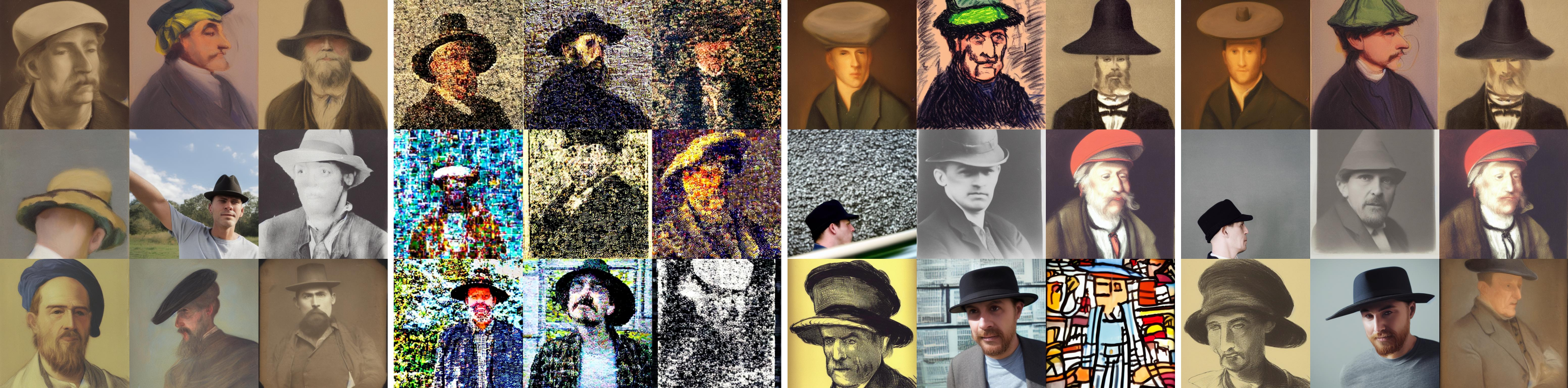}
    \par\centering\vspace{1pt}{\small\textbf{SD1.5}: \emph{A man wearing a hat.}}

    \vspace{2pt}

    \includegraphics[width=\linewidth]{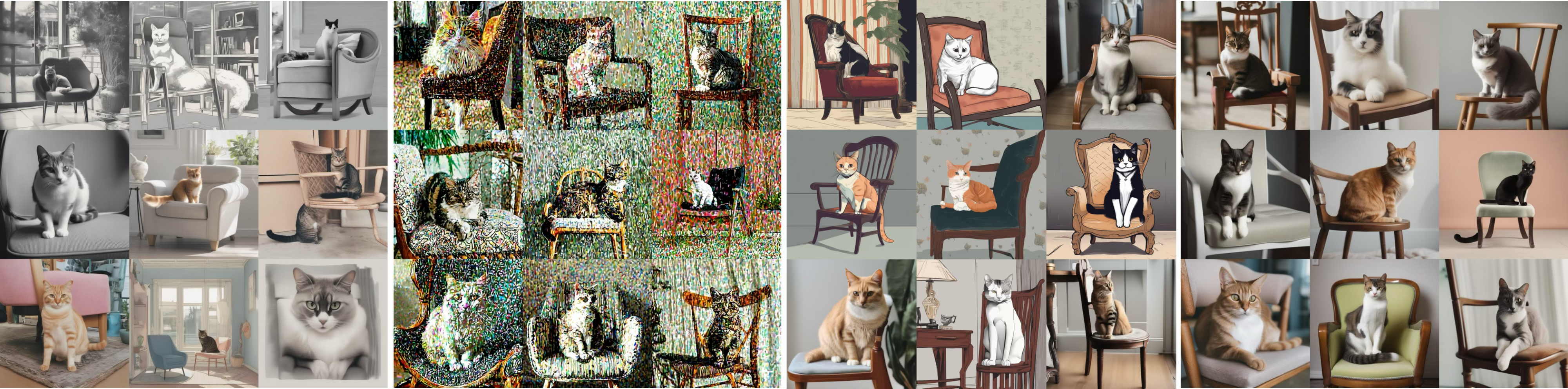}
    \par\centering\vspace{1pt}{\small\textbf{SDXL}: \emph{A cat sitting on a chair.}}

    \vspace{2pt}

    \includegraphics[width=\linewidth]{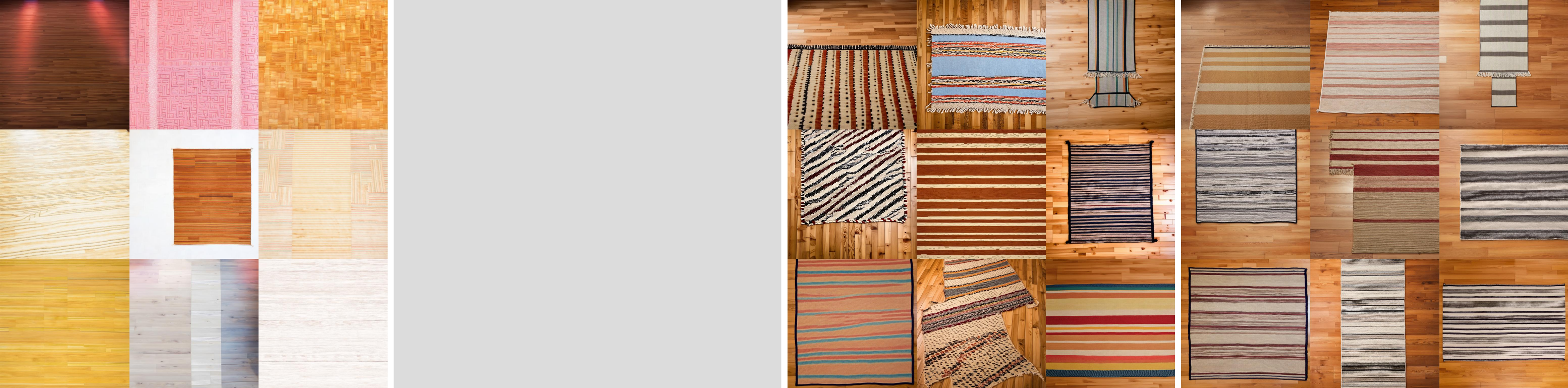}
    \par\centering\vspace{1pt}{\small\textbf{SD3.5}: \emph{The striped rug was on top of the wooden floor.}}

    \vspace{2pt}

    \includegraphics[width=\linewidth]{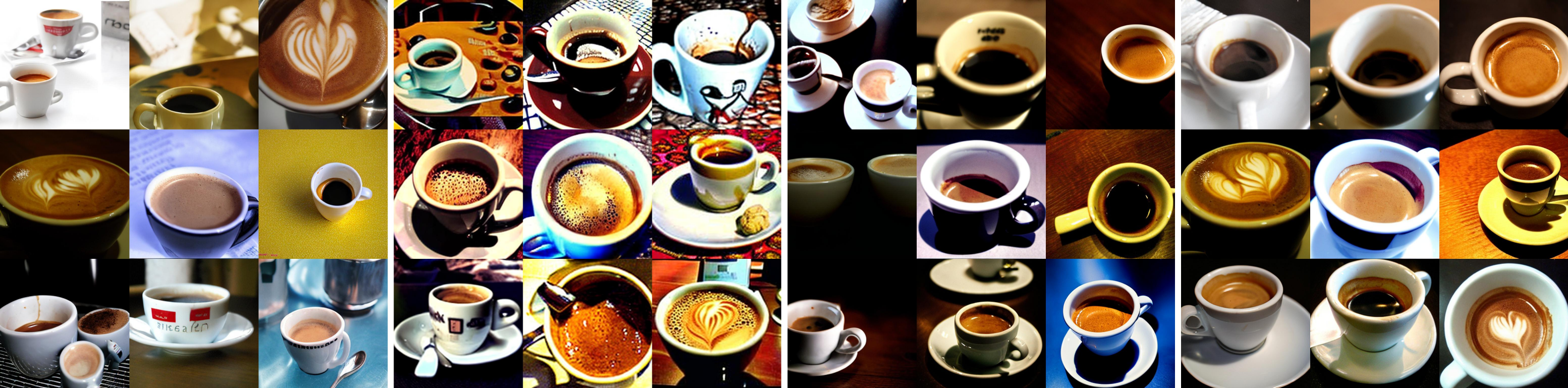}
    \par\centering\vspace{1pt}{\small\textbf{DiT}: \emph{Espresso.}}

    \caption{Qualitative comparison across model architectures. FK is not compatible with flow-matching models and is omitted for SD3.5.}
    \label{fig:comparison-all}
\end{figure*}

%% file: figures/8-comparison-table.tex
\begin{table}[t]
  \caption{\small Quantitative comparison of diversity-enhancement methods.
    CMMD~\cite{jayasumana2024rethinking} is reported for reference (see \Cref{sec:metrics}).
    Error bars are $\pm 2\sigma$.
    Best and second-best entries per model group are \textbf{bold} and \underline{underlined}.}
  \label{tab:comparison}
  \centering
  \small
  \setlength{\tabcolsep}{3pt}
\scalebox{0.8}{
  \begin{tabular}[t]{ll ccc}
    \toprule
    Model & Method & CMMD$\downarrow$ & Vendi$\uparrow$ & Fidelity$\uparrow$ \\
    \midrule
    \multirow{4}{*}{\shortstack[l]{DiT\\$\gamma = 0.20$}}
      & Baseline   & $\mathbf{0.550}$ & $13.45 \pm 0.73$& $\mathbf{0.879 \pm 0.062}$       \\
      & CADS       & $0.753$ & $\mathbf{15.03 \pm 0.34}$& $0.745 \pm 0.094$       \\
      & FK         & $3.138$ & $13.15 \pm 0.99$& $0.745 \pm 0.116$       \\
      & Ours       & $\underline{0.609}$ & $\underline{13.86 \pm 0.74}$& $\underline{0.859 \pm 0.063}$       \\
    \midrule
    \multirow{4}{*}{\shortstack[l]{SD\,1.5\\$\gamma = 0.88$}}
      & Baseline   & $\mathbf{0.546}$ & $14.13 \pm 0.50$& $0.819 \pm 0.058$       \\
      & CADS       & $0.638$ & $\mathbf{14.80 \pm 0.29}$& $\underline{0.820 \pm 0.059}$       \\
      & FK         & $1.249$ & $13.89 \pm 0.58$& $0.817 \pm 0.062$       \\
      & Ours       & $\underline{0.566}$ & $\underline{14.27 \pm 0.49}$& $\mathbf{0.824 \pm 0.056}$       \\
    \bottomrule
    \end{tabular}
    \phantom{A}
    \begin{tabular}[t]{ll ccc}
    \toprule
    Model & Method & CMMD$\downarrow$ & Vendi$\uparrow$ & Fidelity$\uparrow$ \\
    \midrule
    \multirow{4}{*}{\shortstack[l]{SDXL\\$\gamma = 0.94$}}
      & Baseline   & $\underline{0.755}$ & $12.70 \pm 0.68$& $\underline{0.903 \pm 0.035}$       \\
      & CADS       & $0.899$ & $\mathbf{13.89 \pm 0.46}$& $0.810 \pm 0.063$       \\
      & FK         & $1.341$ & $12.94 \pm 0.71$& $0.894 \pm 0.039$    \\
      & Ours       & $\mathbf{0.719}$ & $\underline{13.12 \pm 0.65}$& $\mathbf{0.904 \pm 0.032}$       \\
    \midrule
    \multirow{4}{*}{\shortstack[l]{SD\,3.5\\$\gamma = 0.78$}}
      & Baseline   & $\underline{0.604}$ & $\underline{12.21 \pm 0.86}$& $\mathbf{0.924 \pm 0.036}$       \\
      & CADS      & $1.427$ & $8.29 \pm 0.23$& $0.398 \pm 0.059$       \\
      & FK        & N/A & N/A & N/A \\
      & Ours       & $\mathbf{0.571}$ & $\mathbf{12.43 \pm 0.86}$& $\underline{0.897 \pm 0.056}$       \\
    \bottomrule
  \end{tabular}
} 
\end{table}

%% file: figures/2b-gamma-sweep-avg.tex
\begin{figure}
    \centering
    \includegraphics[width=\linewidth]{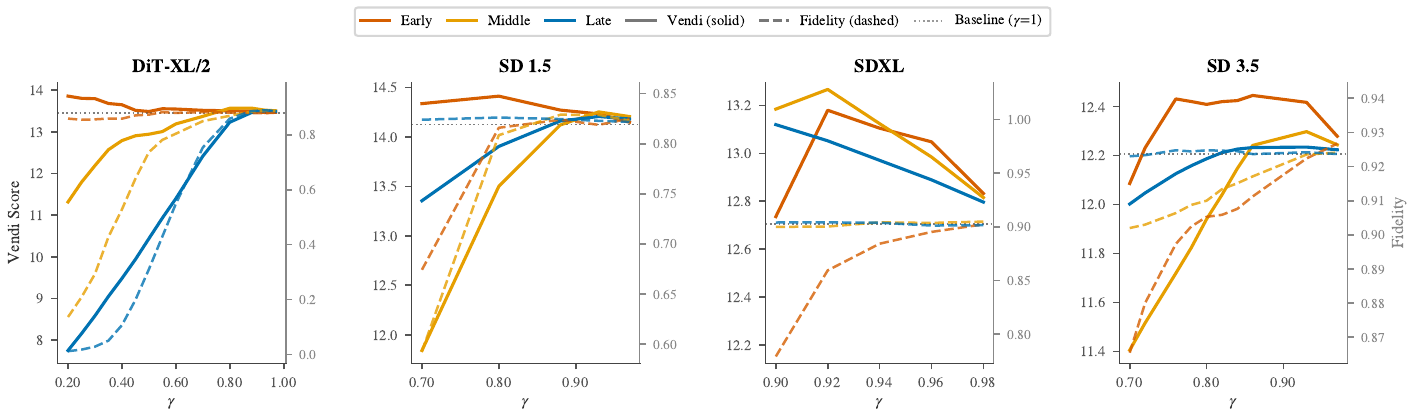}
    \caption{The effect of $\TempParam$ on diversity (Vendi, solid) and fidelity (dashed) for each model, averaged across all condition tiers. 
    }
    \label{fig:gamma-sweep}
\end{figure}

%% file: figures/3-stage-gamma-grid.tex
\begin{figure}
    \centering
    \def\colY{62}
    \def\rowX{-2}
    \vspace{1.5em}
    \begin{overpic}[width=\linewidth]{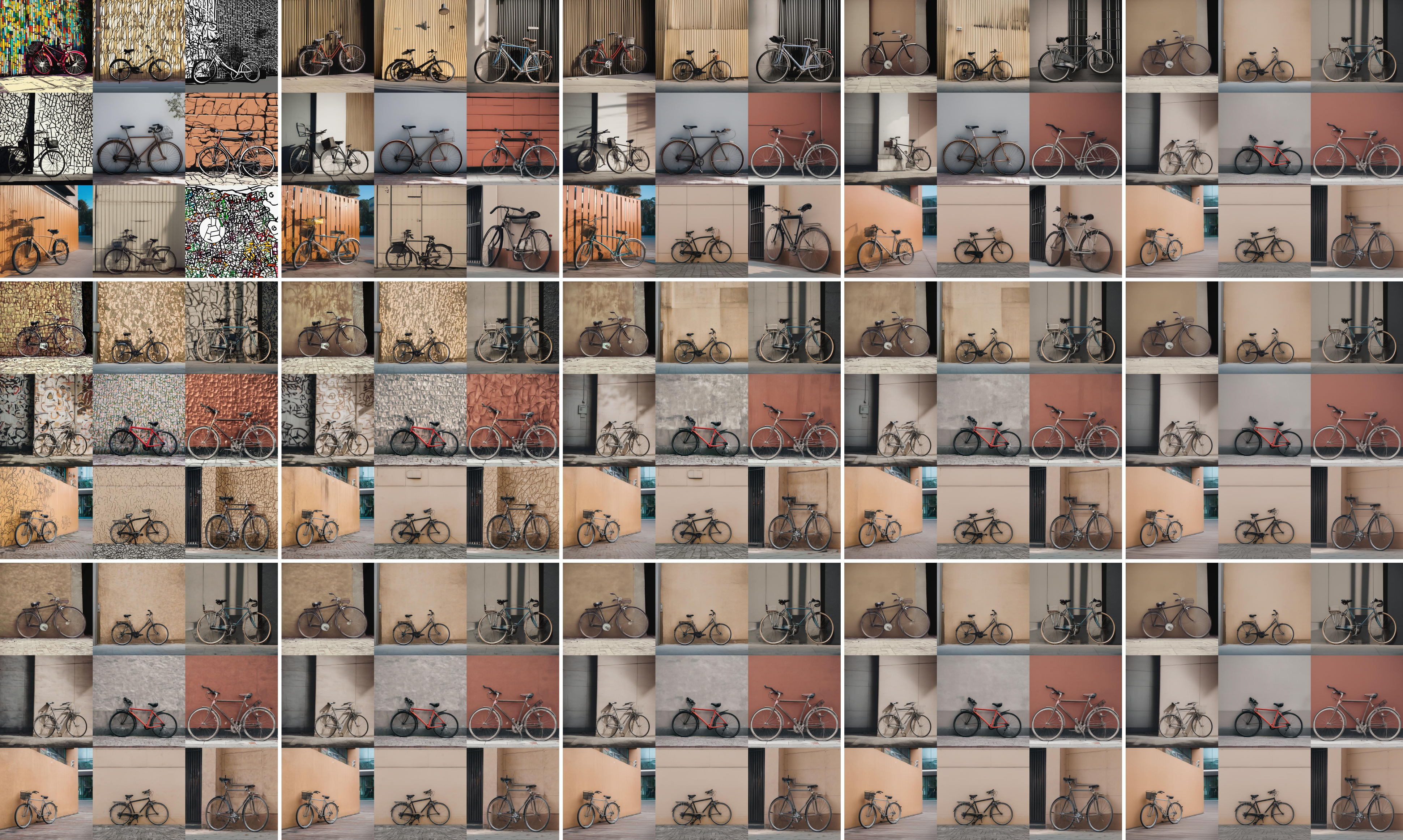}
        \put(10,\colY){\makebox[0pt][c]{\small $\TempParam = 0.90$}}
        \put(30,\colY){\makebox[0pt][c]{\small $\TempParam = 0.92$}}
        \put(50,\colY){\makebox[0pt][c]{\small $\TempParam = 0.94$}}
        \put(70,\colY){\makebox[0pt][c]{\small $\TempParam = 0.96$}}
        \put(90,\colY){\makebox[0pt][c]{\small $\TempParam = 1$ (baseline)}}
        \put(\rowX,47){\makebox[0pt][c]{\rotatebox{90}{\small Early}}}
        \put(\rowX,26.5){\makebox[0pt][c]{\rotatebox{90}{\small Middle}}}
        \put(\rowX,7.5){\makebox[0pt][c]{\rotatebox{90}{\small Late}}}
    \end{overpic}
    \caption{Qualitative effect of $\TempParam$ across denoising stages. 
    Prompt: \emph{A bicycle leaning against a wall} (SDXL).}
    \label{fig:stage-gamma-grid}
\end{figure}

%% file: figures/7-ddim-time-shift.tex
\begin{figure}
    \centering
    \def\colY{77}
    \def\rowX{-3}
    \begin{overpic}[width=\linewidth]{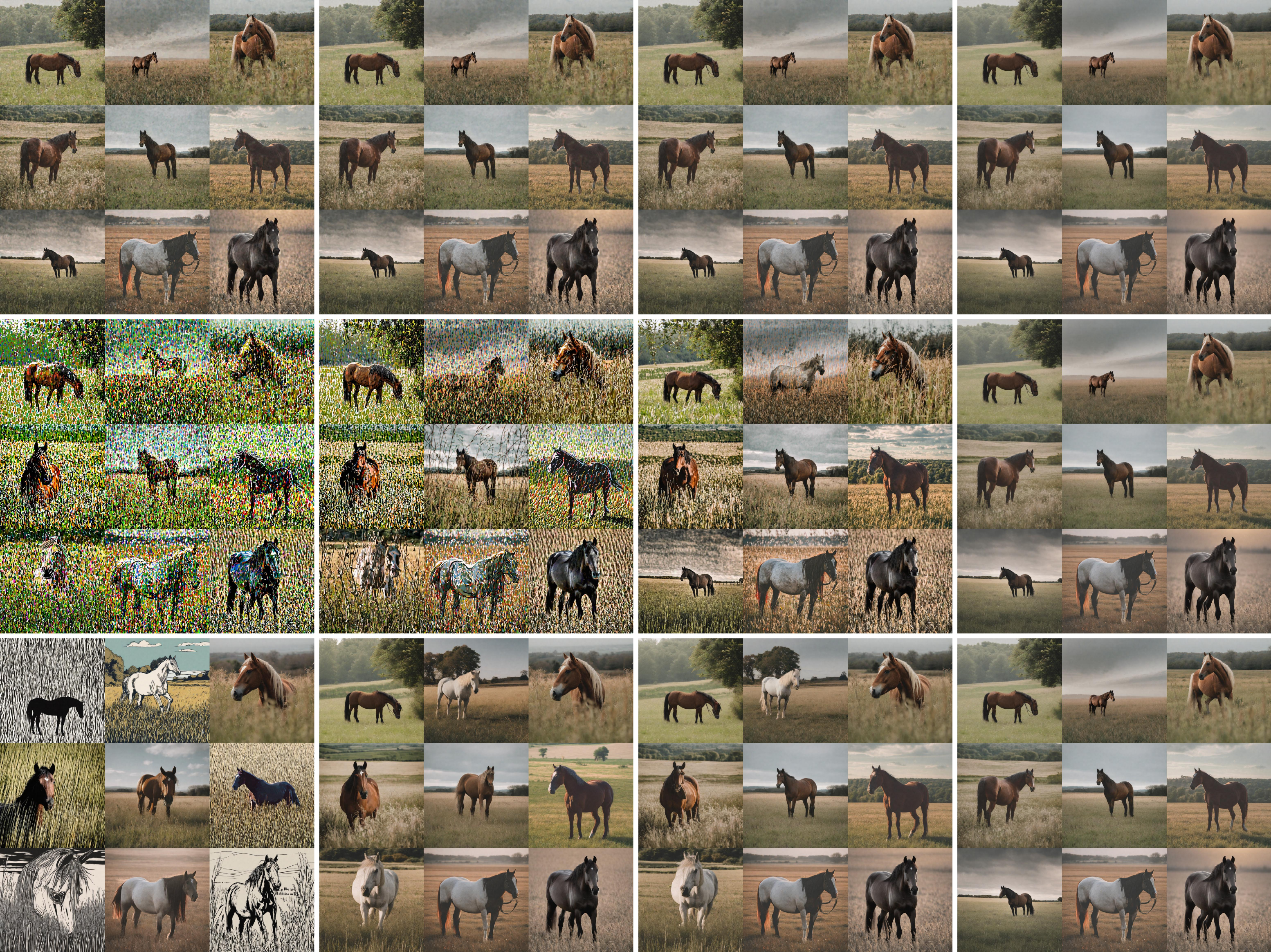}
        \put(12.5,\colY){\makebox[0pt][c]{\small $\TempParam = 0.92$}}
        \put(37.5,\colY){\makebox[0pt][c]{\small $\TempParam = 0.94$}}
        \put(62.5,\colY){\makebox[0pt][c]{\small $\TempParam = 0.96$}}
        \put(87.5,\colY){\makebox[0pt][c]{\small $\TempParam = 1$ (baseline)}}
        \put(\rowX,6.5){\makebox[0pt][c]{\rotatebox{90}{\small Full approach}}}
        \put(\rowX,30.5){\makebox[0pt][c]{\rotatebox{90}{\small Direct tempering}}}
        \put(\rowX,60.5){\makebox[0pt][c]{\rotatebox{90}{\small TSR}}}
    \end{overpic}
    \caption{Effect of variance-corrective time shifting on DDIM across tempering methods and $\TempParam$ values. Prompt: \emph{A horse in a field} (SDXL).}
    \label{fig:ddim-time-shift}
\end{figure}

%% file: 6-conclusion.tex
\section{Discussion}

We have presented variance-corrective time shifting, a training-free method that resolves the variance expansion of naive score scaling and turns temperature sampling into a practical diversity knob for pretrained diffusion and flow-matching models. Studying the effect of $\TempParam$ at different stages of the diffusion process across DiT, Stable Diffusion and Motion Diffusion models on a comprehensive set of conditions, we observe consistent gains in diversity over existing methods at minimal cost to sample quality and condition fidelity.

Our method is bounded by the capacity of the underlying model and cannot overcome its inherent limitations, see, for example, the failure case of a five-legged horse in the bottom-right of the mosaic in \Cref{fig:ddim-time-shift}.
Moreover, temperature scaling with $\TempParam$ is not a free lunch: it trades off fidelity against diversity, and the right balance has to be tuned per model. Our schedule for $\TempParam(t)$ is a three-stage piecewise function obtained by grid search, and a more systematic understanding of how the effect of $\TempParam$ varies along the sampling trajectory together with principled ways to design better schedules are natural directions for future work. 
So is extending the theoretical analysis beyond the toy Gaussian and the well-separated modes assumption to time-varying $\TempParam(t)$ and more complicated settings. 
We are also curious whether the same construction transfers to products of two guidance signals, whether the picture can be connected to entropy and information dynamics~\cite{stancevic2025information} and whether the way $\TempParam$ reshapes generation may in turn reveal something about the structure of the dataset, the conditioning signal and the landscape of the learned distribution.

%% file: 7-appendix.tex
\section{Cross-term negligibility for \Cref{eq:cross-term-negligible}}
\label{app:cross-terms}

The mixture approximation in \Cref{sec:variance-time-shifting} relies on the claim that, under the well-separated mode assumption, the cross terms in $p_t(x)^{\TempParam}$ are negligible for all real $\TempParam > 0$, not just integer values. We make this precise here.

Fix a point $x$ close to center $\mu_k$. Write the mixture as
\begin{equation}
    p_t(x) = f_k(x)\bigl(1 + \delta_k(x)\bigr),
    \qquad
    f_k(x) \coloneqq \pi_k\,\mathcal{N}(x;\mu_k,\sigma_t^2 I),
\end{equation}
where $\delta_k(x)$ collects the contributions from all other components:
\begin{equation}
    \delta_k(x) \coloneqq \sum_{j \neq k} \frac{\pi_j\,\mathcal{N}(x;\mu_j,\sigma_t^2 I)}{\pi_k\,\mathcal{N}(x;\mu_k,\sigma_t^2 I)}.
\end{equation}

Expanding $\|x - \mu_j\|^2$ around $\mu_k$, the per-component Gaussian ratio reads
\begin{equation}
    \frac{\mathcal{N}(x;\mu_j,\sigma_t^2 I)}{\mathcal{N}(x;\mu_k,\sigma_t^2 I)}
    = \exp\!\left(-\frac{\|\mu_k - \mu_j\|^2 + 2(x-\mu_k)^\top(\mu_k - \mu_j)}{2\sigma_t^2}\right).
    \label{eq:gauss-ratio}
\end{equation}
Restricting to the support of mode $k$, namely $\|x - \mu_k\| \le r$ for some $r$ on the order of $\sigma_t$, the Cauchy--Schwarz inequality bounds the linear term by $|2(x-\mu_k)^\top(\mu_k - \mu_j)| \le 2r\,\|\mu_k - \mu_j\|$, so \Cref{eq:gauss-ratio} yields the uniform estimate
\begin{equation}
    \frac{\mathcal{N}(x;\mu_j,\sigma_t^2 I)}{\mathcal{N}(x;\mu_k,\sigma_t^2 I)}
    \le \exp\!\left(-\frac{\|\mu_k - \mu_j\|\bigl(\|\mu_k - \mu_j\| - 2r\bigr)}{2\sigma_t^2}\right).
\end{equation}
Under the well-separated assumption $\|\mu_k - \mu_j\| \gg \max(\sigma_t,\,2r)$, every summand of $\delta_k(x)$ is exponentially small, so $\delta_k(x) \ll 1$ uniformly over the support of mode $k$.

For any real $\TempParam > 0$ and $|\delta| < 1$, the generalized binomial series gives
\begin{equation}
    (1 + \delta)^{\TempParam} = \sum_{n=0}^\infty \binom{\TempParam}{n}\delta^n
    = 1 + \TempParam\,\delta + \frac{\TempParam(\TempParam-1)}{2}\delta^2 + \cdots
    \label{eq:gen-binom}
\end{equation}
Applying \Cref{eq:gen-binom} to $p_t(x)^{\TempParam} = f_k(x)^{\TempParam}(1+\delta_k)^{\TempParam}$ gives
\begin{equation}
    p_t(x)^{\TempParam}
    = f_k(x)^{\TempParam}\!\left[1 + \TempParam\,\delta_k + O(\delta_k^2)\right].
\end{equation}
Because $\delta_k = O\!\left(\exp(-\|\mu_k - \mu_j\|^2/2\sigma_t^2)\right)$ is exponentially small, all correction terms are negligible and
\begin{equation}
    p_t(x)^{\TempParam} \approx f_k(x)^{\TempParam} = \pi_k^{\TempParam}\,\mathcal{N}(x;\mu_k,\sigma_t^2 I)^{\TempParam}.
\end{equation}

Since the mode supports are disjoint under the well-separated assumption, summing the local approximations gives
\begin{equation}
    p_t(x)^{\TempParam} \approx \sum_k \pi_k^{\TempParam}\,\mathcal{N}(x;\mu_k,\sigma_t^2 I)^{\TempParam}.
\end{equation}
This holds for all real $\TempParam > 0$. The further proportionality
\begin{equation}
    \mathcal{N}(x;\mu_k,\sigma_t^2 I)^{\TempParam}
    \propto \exp\!\left(-\frac{\TempParam\,\|x-\mu_k\|^2}{2\sigma_t^2}\right)
    \propto \mathcal{N}\!\left(x;\mu_k,\tfrac{\sigma_t^2}{\TempParam}I\right)
\end{equation}
follows by inspection of the Gaussian density.

\section{Temperature sampling for variance-preserving diffusion models}
\label{sec:appendix-vp}

The main text motivates variance-corrective time shifting for temperature sampling. This appendix focuses on the VP case and, in particular, how to implement the time shift with an $\epsilon$-predicting network.

\subsection{Key identities}

We use the standard VP parameterization
\begin{equation}
	x_t = \sqrt{\bar\alpha_t}\,x_0 + \sqrt{1-\bar\alpha_t}\,\epsilon,
	\qquad
	\epsilon\sim\mathcal{N}(0,I),
\end{equation}
so that an $\epsilon$-predictor yields the score estimate
\begin{equation}
	\nabla_x \log p_t(x) \approx -\frac{\epsilon_\theta(x,t)}{\sqrt{1-\bar\alpha_t}}.
\end{equation}
It is convenient to work with the VE-equivalent noise scale
\begin{equation}
	\sigma_t^2 \coloneqq \frac{1-\bar\alpha_t}{\bar\alpha_t},
\end{equation}
which is monotone in $t$.

Time shifting defines a shifted time $\tilde t$ by matching the effective noise level in the normalized variable $u_t = x_t/\sqrt{\bar\alpha_t}$, i.e.
\begin{equation}
	\sigma_{\tilde t}^2 = \TempParam\,\sigma_t^2
	\qquad
	(\text{equivalently, }\sigma_{\tilde t}=\sqrt{\TempParam}\,\sigma_t),
\end{equation}
or, in terms of $\bar\alpha_t$,
\begin{equation}
	\bar\alpha_{\tilde t} = \frac{\bar\alpha_t}{\bar\alpha_t + \TempParam\,(1-\bar\alpha_t)}.
\end{equation}

\subsection{Practical score implementation}

Assume access to a pre-trained VP noise predictor $\epsilon_\theta(x_t,t)$ (standard $\epsilon$-prediction, $\epsilon\sim\mathcal{N}(0,I)$). The temperature-scaled score follows directly from the VE result \eqref{eq:time-shifted-score} by working with the normalized variable $u_t \coloneqq x_t/\sqrt{\bar\alpha_t}$, in which the VP forward kernel reduces to a pure VE kernel $\mathcal{N}(x_0,\sigma_t^2 I)$ with $\sigma_t^2 = (1-\bar\alpha_t)/\bar\alpha_t$. Writing $p^u_t$ for the marginal of $u_t$ and $p^{u,(\TempParam)}_t$ for its tempered counterpart, the argument of \Cref{eq:time-shifted-score} transports verbatim into $u$-space,
\begin{equation}
	\nabla_u \log p^{u,(\TempParam)}_t(u) \;\approx\; \TempParam \cdot \nabla_u \log p^u_{\tilde t}(u),
	\label{eq:vp-u-temp-score}
\end{equation}
with the shifted timestep $\tilde t$ chosen by nearest-neighbor lookup on the (monotone) schedule $\{\sigma_t\}_t$ to match the target noise level $\sigma_{\tilde t} = \sqrt{\TempParam}\,\sigma_t$,
\begin{equation}
	\tilde t \in \argmin_{t'}\,\bigl|\sigma_{t'} - \sqrt{\TempParam}\,\sigma_t\bigr|.
\end{equation}
Since the VP network takes $x$ (not $u$) as input, switching from $t$ to $\tilde t$ while keeping $u$ fixed requires rescaling the input,
\begin{equation}
	\tilde x \coloneqq \sqrt{\bar\alpha_{\tilde t}}\,u \;=\; \sqrt{\frac{\bar\alpha_{\tilde t}}{\bar\alpha_t}}\,x,
\end{equation}
so that the network call resolves to $\epsilon_\theta(\tilde x,\tilde t)$. Converting the $u$-space score \eqref{eq:vp-u-temp-score} back to $x$-space at the original time $t$ via $x = \sqrt{\bar\alpha_t}\,u$ contributes a Jacobian $1/\sqrt{\bar\alpha_t}$, which yields the VE-parallel form
\begin{equation}
	\nabla_x \log \TempDistT{t}(x) \;\approx\; \TempParam \cdot \left(-\frac{\epsilon_\theta(\tilde x,\tilde t)}{\sqrt{\bar\alpha_t}\,\sigma_{\tilde t}}\right).
	\label{eq:vp-temp-explicit}
\end{equation}
At $\TempParam=1$, $\sqrt{\bar\alpha_t}\,\sigma_t = \sqrt{1-\bar\alpha_t}$ and the bracketed expression reduces to the standard VP score conversion $-\epsilon_\theta(x,t)/\sqrt{1-\bar\alpha_t}$, as expected.

The two variance factors collapse algebraically. Using $\sigma_{\tilde t}^2 = \TempParam\,(1-\bar\alpha_t)/\bar\alpha_t$,
\begin{equation}
	\sqrt{\bar\alpha_t}\,\sigma_{\tilde t} \;=\; \sqrt{\bar\alpha_t \cdot \TempParam\,(1-\bar\alpha_t)/\bar\alpha_t} \;=\; \sqrt{\TempParam}\,\sqrt{1-\bar\alpha_t},
\end{equation}
so one factor of $\sqrt{\TempParam}$ in the numerator of \Cref{eq:vp-temp-explicit} cancels with the variance factor in the denominator and the expression simplifies to
\begin{equation}
	\nabla_x \log \TempDistT{t}(x) \;\approx\; -\frac{\sqrt{\TempParam}\,\epsilon_\theta(\tilde x,\tilde t)}{\sqrt{1-\bar\alpha_t}}.
	\label{eq:vp-temp-compact}
\end{equation}
The compact form \eqref{eq:vp-temp-compact} matches the standard VP score conversion at the original time $t$ with the network query relocated to $(\tilde x,\tilde t)$ and a single $\sqrt{\TempParam}$ scalar applied to the output. Operationally, this can be implemented as a wrapper that returns
\begin{equation}
	\epsilon_{\text{temp}}(x,t) \;\coloneqq\; \sqrt{\TempParam}\,\epsilon_\theta(\tilde x,\tilde t)
\end{equation}
in place of $\epsilon_\theta(x,t)$, leaving the VP sampler and time schedule unchanged at the original timestep $t$.

\section{Temperature sampling for flow-matching models}
\label{app:flow-matching}

Modern open-weight text-to-image backbones such as Stable Diffusion 3~\cite{esser2024scaling} are trained as flow-matching models~\cite{lipman2022flow, liu2022flow} rather than VE/VP diffusion. Variance-corrective time shifting still applies in this setting with no retraining, because a rectified-flow forward kernel is just another affine Gaussian and the analysis of \Cref{eq:time-shifted-score} carries through after a change of variables.

\subsection{Key identities}

We use the rectified-flow parameterization
\begin{equation}
	x_t = (1-t)\,x_0 + t\,\epsilon,
	\qquad \epsilon \sim \mathcal{N}(0,I),\quad t \in [0,1],
\end{equation}
so $t=0$ is data and $t=1$ is noise. The conditional kernel $p(x_t\mid x_0) = \mathcal{N}(x_t;(1-t)x_0,\,t^2 I)$ is affine with signal scale $a_t = 1-t$ and noise scale $b_t = t$. The model predicts the velocity
\begin{equation}
	v_\theta(x_t,t) \approx \mathbb{E}[\epsilon - x_0 \mid x_t],
\end{equation}
from which the noise prediction and the score follow algebraically:
\begin{align}
	\epsilon_\theta(x,t) &= (1-t)\,v_\theta(x,t) + x,
	\label{eq:fm-eps-from-v}\\
	\nabla_x \log p_t(x) &\approx -\frac{\epsilon_\theta(x,t)}{t}
	= -\frac{(1-t)\,v_\theta(x,t) + x}{t}.
	\label{eq:fm-score-from-v}
\end{align}
Defining the normalized variable $u_t \coloneqq x_t / a_t = x_t/(1-t)$ reduces the conditional to a pure VE form, $u_t \mid x_0 \sim \mathcal{N}(x_0,\,\sigma_e(t)^2 I)$, with effective noise scale
\begin{equation}
	\sigma_e(t) \;\coloneqq\; \frac{b_t}{a_t} \;=\; \frac{t}{1-t}.
\end{equation}

\subsection{Time-shift for rectified flow}

By the same argument as \Cref{eq:time-shifted-score}, the tempered marginal in $u$-space is approximated by direct tempering of $p^u_{\tilde t}$ provided $\sigma_e(\tilde t)^2 = \TempParam\,\sigma_e(t)^2$. For the rectified-flow schedule this gives
\begin{equation}
	\frac{\tilde t}{1-\tilde t} \;=\; \sqrt{\TempParam}\,\frac{t}{1-t}
	\qquad\Longleftrightarrow\qquad
	\tilde t \;=\; \frac{\sqrt{\TempParam}\,t}{1 - (1-\sqrt{\TempParam})\,t}.
	\label{eq:fm-t-shift}
\end{equation}
For $\TempParam < 1$ this yields $\tilde t < t$, i.e.\ a query at a slightly less noisy time. The corresponding input rescaling that keeps the normalized variable $u$ fixed across the time shift is
\begin{equation}
	\tilde x \;\coloneqq\; \frac{a_{\tilde t}}{a_t}\,x \;=\; \frac{1-\tilde t}{1-t}\,x \;=\; \frac{x}{1 - (1-\sqrt{\TempParam})\,t}.
	\label{eq:fm-x-shift}
\end{equation}

\subsection{Practical velocity implementation}

As in the VP case, the temperature-scaled score is most transparently derived in $u$-space, where the rectified-flow kernel reduces to VE with noise scale $\sigma_e(t)=t/(1-t)$. Applying \Cref{eq:time-shifted-score} at the shifted noise level $\sigma_e(\tilde t) = \sqrt{\TempParam}\,\sigma_e(t)$ gives
\begin{equation}
	\nabla_u \log p^{u,(\TempParam)}_t(u) \;\approx\; \TempParam \cdot \nabla_u \log p^u_{\tilde t}(u),
\end{equation}
where $p^u_t$ is the marginal of the normalized variable $u_t = x_t/(1-t)$ introduced in the previous subsection, and $p^{u,(\TempParam)}_t$ is its tempered counterpart. The input rescaling \eqref{eq:fm-x-shift} ensures the network call resolves to $\epsilon_\theta(\tilde x,\tilde t)$, and converting the score back to $x$-space at the original time $t$ via $x = (1-t)\,u$ contributes a Jacobian $1/(1-t)$, yielding the VE-parallel form
\begin{equation}
	\nabla_x \log \TempDistT{t}(x) \;\approx\; \TempParam \cdot \left(-\frac{\epsilon_\theta(\tilde x,\tilde t)}{(1-t)\,\sigma_e(\tilde t)}\right).
	\label{eq:fm-temp-explicit}
\end{equation}
Using $\sigma_e(\tilde t) = \sqrt{\TempParam}\,t/(1-t)$, the variance factor collapses,
\begin{equation}
	(1-t)\,\sigma_e(\tilde t) \;=\; \sqrt{\TempParam}\,t,
\end{equation}
so one $\sqrt{\TempParam}$ cancels and \Cref{eq:fm-temp-explicit} simplifies to $-\sqrt{\TempParam}\,\epsilon_\theta(\tilde x,\tilde t)/t$. The corresponding tempered noise prediction at the original time is therefore
\begin{equation}
	\epsilon_{\text{temp}}(x,t)
	\;\coloneqq\; \sqrt{\TempParam}\,\epsilon_\theta(\tilde x,\tilde t),
\end{equation}
identical in structure to the VP case. Composing with \Cref{eq:fm-eps-from-v},
\begin{equation}
	\epsilon_{\text{temp}}(x,t)
	\;=\; \sqrt{\TempParam}\,\bigl[(1-\tilde t)\,v_\theta(\tilde x,\tilde t) + \tilde x\bigr],
\end{equation}
and converting back to a velocity at the original time $t$ via $v = (\epsilon - x)/(1-t)$ yields the tempered velocity to plug into the unmodified flow-matching sampler:
\begin{equation}
	\boxed{\;
	v_{\text{temp}}(x,t)
	\;=\; \frac{\sqrt{\TempParam}\,\bigl[(1-\tilde t)\,v_\theta(\tilde x,\tilde t) + \tilde x\bigr] - x}{1-t}
	\;}
	\label{eq:fm-v-temp}
\end{equation}
with $\tilde t$ and $\tilde x$ given by \Cref{eq:fm-t-shift,eq:fm-x-shift}. At $\TempParam = 1$ we recover $\tilde t = t$, $\tilde x = x$ and $v_{\text{temp}} = v_\theta$, as expected. 

\section{Vendi score}
\label{app:vendi}

The Vendi score~\cite{friedman2022vendi} measures the effective number of distinct modes in a finite sample. Given $n$ samples $x_1,\dots,x_n \in \mathcal{X}$ and a symmetric positive-semidefinite similarity kernel $k: \mathcal{X}\times\mathcal{X}\to[0,1]$ satisfying $k(x,x)=1$, form the $n\times n$ Gram matrix $K$ with $K_{ij} = k(x_i,x_j)$ and normalize it to $\hat{K} = K/n$. Let $\lambda_1,\dots,\lambda_n \geq 0$ be the eigenvalues of $\hat{K}$, which satisfy $\sum_i \lambda_i = 1$ by construction. The Vendi score is
\begin{equation}
    \mathrm{VS}(x_1,\dots,x_n)
    \;\coloneqq\;
    \exp\!\left(-\sum_{i=1}^{n} \lambda_i \log \lambda_i\right)
    \;=\;
    \exp\left(H(\hat{K})\right),
\end{equation}
where $H(\hat{K}) = -\operatorname{tr}(\hat{K}\log\hat{K})$ is the von Neumann entropy of $\hat{K}$. The score equals 1 when all samples are identical (rank-one $\hat{K}$) and equals $n$ when all samples are mutually orthogonal under $k$ (identity $\hat{K}$), making it an interpretable effective count of distinct elements.

For image samples we follow CADS~\cite{sadat2024cads} and use cosine similarity on SSCD~\cite{pizzi2022self} features as the kernel: $k(x_i,x_j) = \phi(x_i)^\top\phi(x_j) / (\|\phi(x_i)\|\|\phi(x_j)\|)$, where $\phi$ is the SSCD encoder. SSCD features are designed to be invariant to photometric variation while remaining sensitive to semantic and compositional differences, making them well-suited for measuring diversity beyond low-level pixel statistics.

\section{Ablation setup and conditioning sets}
\label{app:prompts}

We organize the ablations around a per-condition sweep. For each configuration we draw $K = 16$ samples per condition and report the Vendi score averaged over conditions within a prompt tier. Smaller $\TempParam$ values were excluded after preliminary runs showed unrecoverable loss of prompt fidelity. The diffusion trajectory is partitioned into three contiguous stages of equal timestep length (early, mid and late) and $\TempParam$ is applied within a single stage per run, with $\TempParam$ applied across the full trajectory included as an additional configuration. The untempered model at $\TempParam = 1$ provides the baseline.

The following subsections enumerate the full conditioning sets. For text-conditioned models we use 40 prompts split across three tiers of decreasing diversity headroom: 15 single-concept prompts (L), 15 mildly descriptive prompts (M) and 10 compositional prompts with explicit attributes, counts or spatial relations drawn from T2I-CompBench~\cite{huang2023t2i}. For ImageNet class-conditional models we use 20 classes evenly split between a high-variance subset (scenes and broad concepts) and a canonical subset (objects with stereotyped appearance).

\subsection{Text-to-image prompts}

\paragraph{Tier L: single-concept prompts (15).}
\begin{itemize}\itemsep0pt
\item a dog
\item a landscape
\item a portrait of a person
\item a building
\item a bowl of food
\item a vehicle on a road
\item a mountain
\item an oil painting
\item a still life
\item a child playing
\item a city street
\item a flower
\item a bird in flight
\item a piece of furniture
\item a dessert
\end{itemize}

\paragraph{Tier M: mildly descriptive prompts (15).}
\begin{itemize}\itemsep0pt
\item a cat sitting on a chair
\item a man wearing a hat
\item a forest in autumn
\item a car parked at night
\item a bowl of soup on a table
\item a woman reading a book
\item a dog running on a beach
\item a kitchen with morning light
\item a bicycle leaning against a wall
\item a horse in a field
\item a coffee shop interior
\item a sunset over water
\item a chef preparing a meal
\item a vase of flowers on a windowsill
\item a bookshop with old books
\end{itemize}

\paragraph{Tier T: compositional prompts (10).}
Drawn from T2I-CompBench~\cite{huang2023t2i}.

\begin{itemize}\itemsep0pt
	\item the red hat was on top of the brown coat rack.
	\item the fluffy cat is on the left of the soft pillow.
	\item the rectangular picture frame was hung above the beige couch.
	\item the leather wallet was inside the brown purse.
	\item the black chair was on the left of the white table.
	\item the red apple was on top of the black plate.
	\item the striped rug was on top of the wooden floor.
	\item the silver laptop was on top of the black desk.
	\item the square box was next to the circular canister.
	\item the rough brick was on top of the smooth tile.
\end{itemize}

\subsection{ImageNet classes}

\paragraph{Subset H: high intra-class variance (10).}
Scenes and broad categories.
\begin{itemize}\itemsep0pt
\item seashore
\item valley
\item lakeside
\item alp
\item library
\item restaurant
\item jigsaw puzzle
\item comic book
\item bookshop
\item toyshop
\end{itemize}

\paragraph{Subset C: canonical low-variance (10).}
Objects with stereotyped pose or appearance.
\begin{itemize}\itemsep0pt
\item great white shark
\item hammerhead shark
\item golf ball
\item balance beam
\item espresso
\item pretzel
\item soccer ball
\item volleyball
\item screw
\item spotlight
\end{itemize}

\input{figures/2-gamma-sweep}

\Cref{fig:gamma-sweep-full} reports the per-tier breakdown of the $\TempParam$ sweep summarized in \Cref{fig:gamma-sweep} of the main text. The trends from the averaged view carry over to each tier. Two additional observations emerge from the breakdown. First, at $\TempParam = 1$ the loose tier attains higher diversity than the tight tier, confirming that the tier partition reflects a real gap in diversity headroom. Second, the diversity gain from our method is more pronounced on the tight tier than on the loose tier, consistent with the expectation that prompts with tighter constraints have more room to improve.

\section{Qualitative evaluation on Motion Diffusion models}
\label{app:motion-diffusion}

\input{figures/9-comparison-human.tex}

As shown in \Cref{fig:comparison-human}, CADS noticeably degrades condition fidelity across all four prompts. Our method produces more diverse motions while remaining faithful to the text prompt. We encourage readers to view the accompanying video for a fuller comparison.

\section{Computational resources}
\label{app:compute}

We do not train any models for this paper. All experiments are conducted with off-the-shelf checkpoints that are publicly available and widely used in the community. We generate around 100k samples across all experiments used in the paper, which we estimate to require around 650 GPU-hours on an nVidia RTX 3090. The Vendi score is computed on CPU and takes around 10 hours for the full set of samples.

%% file: figures/2-gamma-sweep.tex
\begin{figure*}[h]
    \centering
    \includegraphics[width=\linewidth]{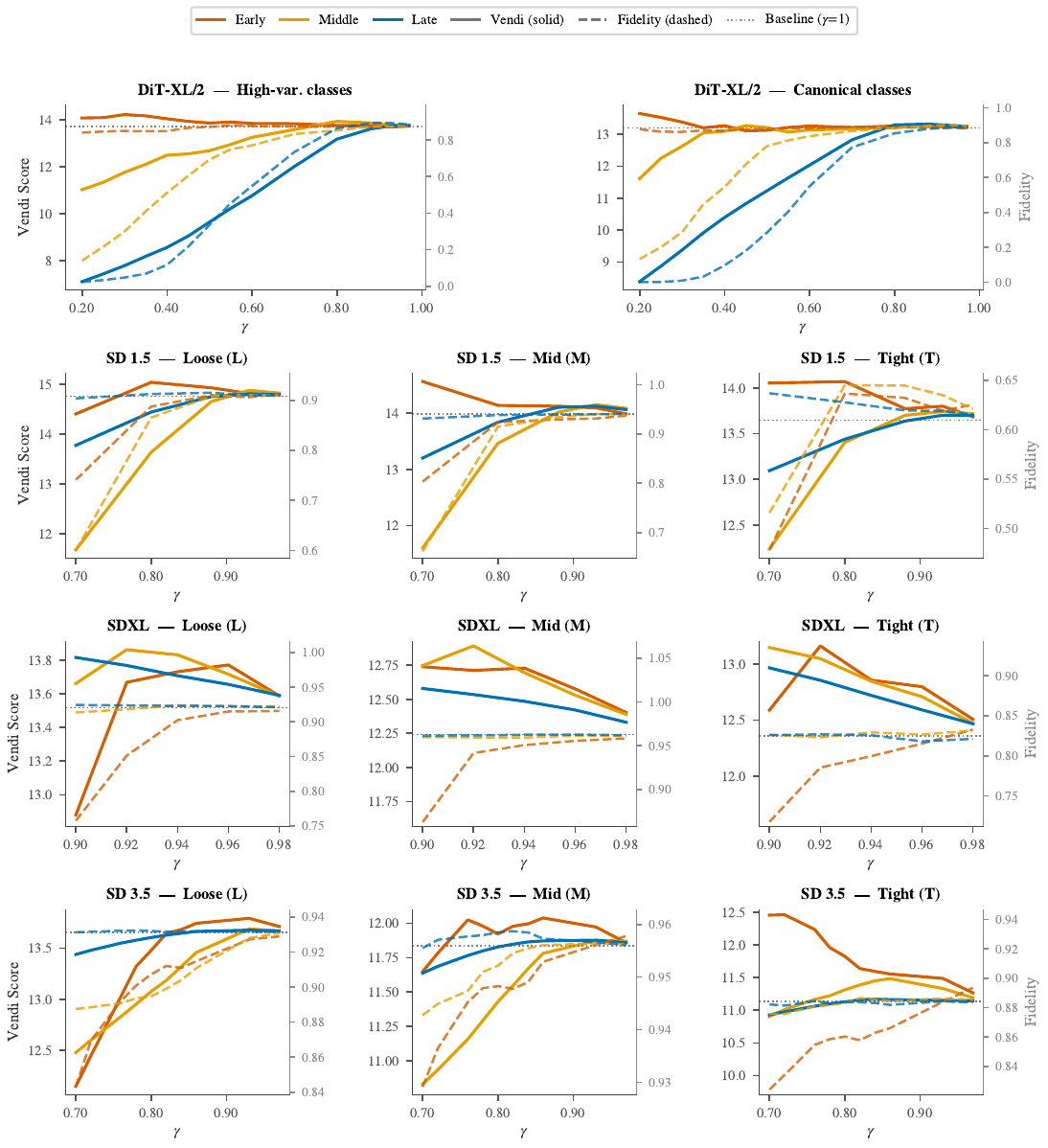}
    \caption{Per-tier breakdown of the $\TempParam$ sweep (full version of \Cref{fig:gamma-sweep}). Each row corresponds to one model and each column corresponds to one condition tier (prompt tier for text-conditioned models, class subset for ImageNet). Within each panel, $\TempParam$ is on the horizontal axis,  and we overlay six curves: three denoising stages (early, middle, late) crossed with two metrics, Vendi (diversity, left axis) and VQA/Accuracy (fidelity, right axis). The dotted line marks the untempered baseline ($\TempParam = 1$).}
    \label{fig:gamma-sweep-full}
\end{figure*}

%% file: figures/9-comparison-human.tex
\begin{figure*}
    \centering

    \makebox[0.33\linewidth][c]{\small CADS}%
    \makebox[0.33\linewidth][c]{\small Ours}%
    \makebox[0.33\linewidth][c]{\small Baseline ($\gamma = 1$)}

    \vspace{3pt}

    \includegraphics[width=\linewidth]{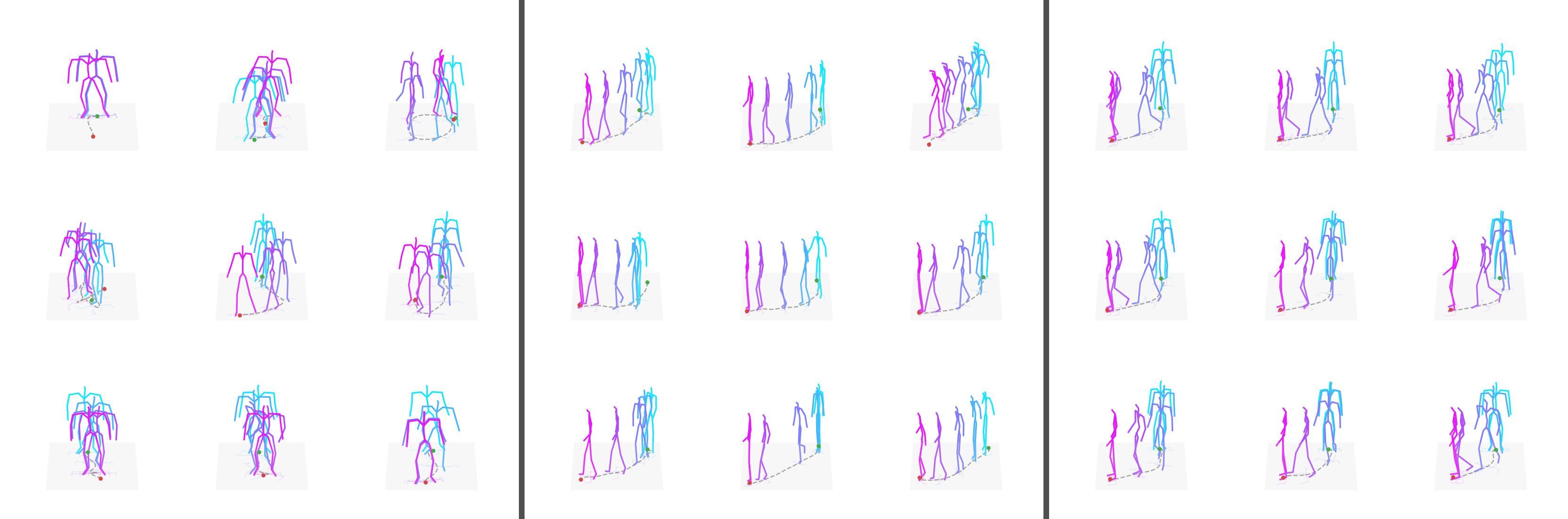}
    \par\centering\vspace{1pt}{\small\emph{A figure walks forward and to the right}}

    \vspace{3pt}

    \includegraphics[width=\linewidth]{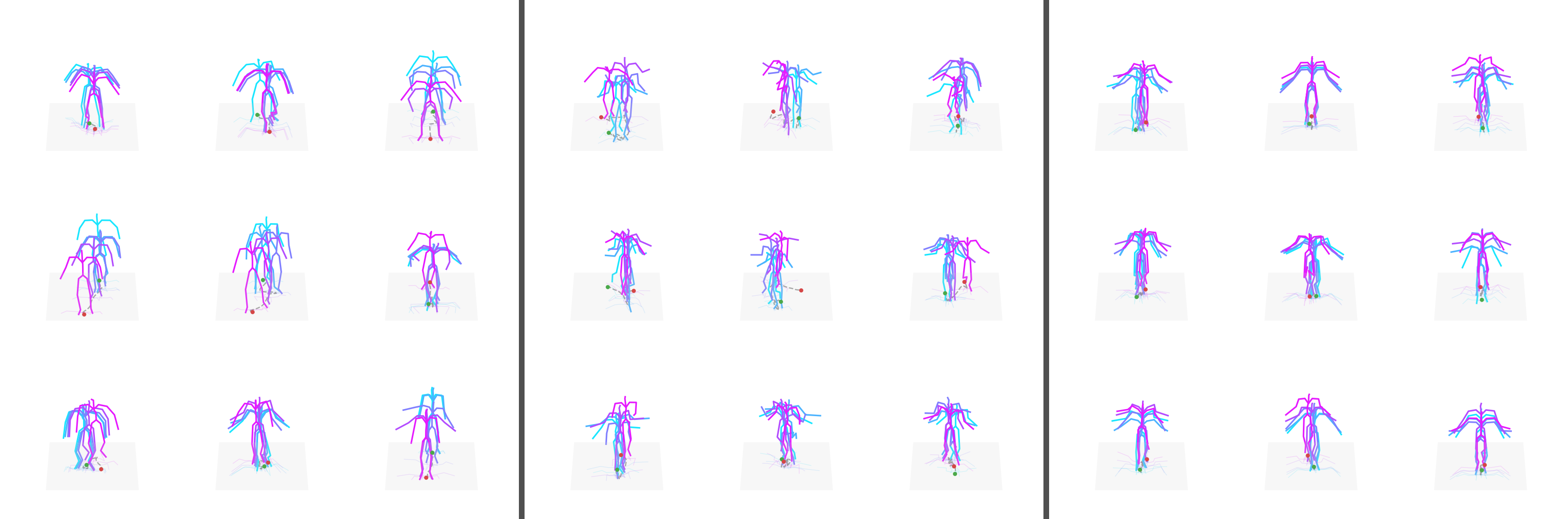}
    \par\centering\vspace{1pt}{\small\emph{A person is skipping rope}}

    \vspace{3pt}

    \includegraphics[width=\linewidth]{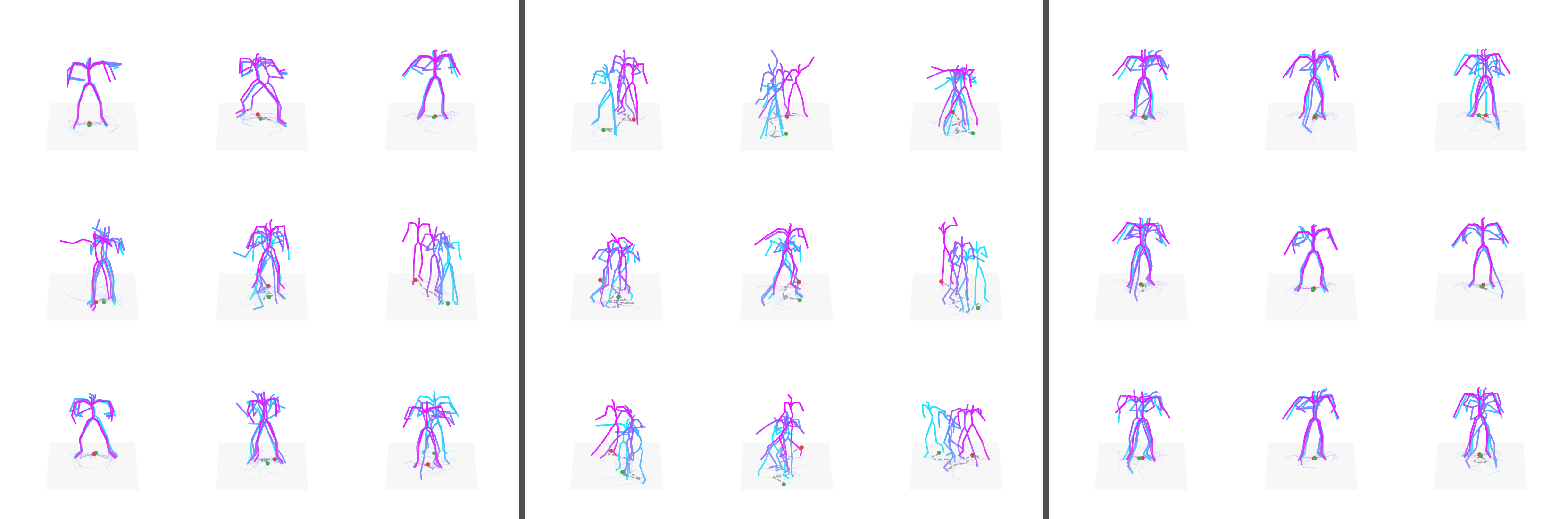}
    \par\centering\vspace{1pt}{\small\emph{A person punches in a manner consistent with martial arts}}

    \vspace{4pt}

    \includegraphics[width=\linewidth]{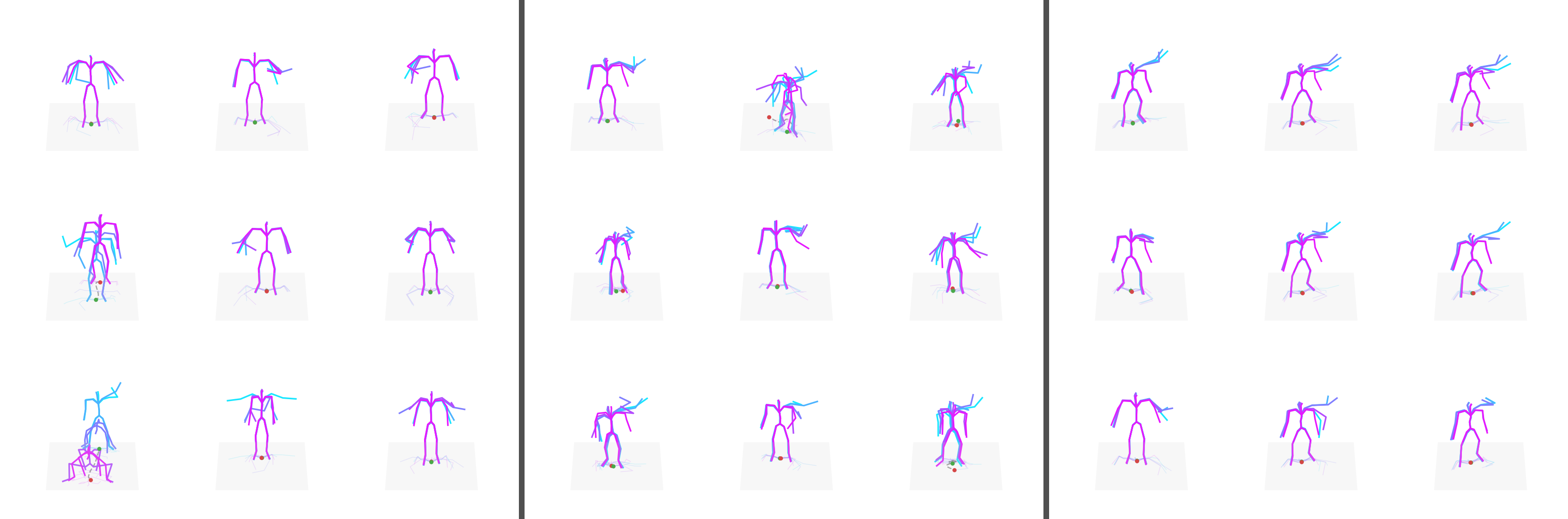}
    \par\centering\vspace{1pt}{\small\emph{The person is waving their left arm}}

    \caption{Qualitative comparison on the motion diffusion model~\cite{tevet2022human}. Each row shows one text prompt, and each column shows samples from CADS, our method and the baseline ($\TempParam=1$). Five evenly spaced frames are rendered as skeletons, with color encoding time from cyan (earliest) to magenta (latest). A dashed floor trajectory traces the root joint (pelvis) path over the full sequence.}
    \label{fig:comparison-human}
\end{figure*}

%% file: bibs.bib
@inproceedings{sadat2024cads,
  author       = {Seyedmorteza Sadat and
                  Jakob Buhmann and
                  Derek Bradley and
                  Otmar Hilliges and
                  Romann M. Weber},
  title        = {{CADS:} Unleashing the Diversity of Diffusion Models through Condition-Annealed
                  Sampling},
  booktitle    = {The Twelfth International Conference on Learning Representations,
                  {ICLR} 2024, Vienna, Austria, May 7-11, 2024},
  year         = {2024},
}

@article{xu2025temporal,
  title={Temporal Score Rescaling for Temperature Sampling in Diffusion and Flow Models},
  author={Xu, Yanbo and Wu, Yu and Park, Sungjae and Zhou, Zhizhuo and Tulsiani, Shubham},
  journal={arXiv preprint arXiv:2510.01184},
  year={2025}
}

@article{pandey2024heavy,
  title={Heavy-tailed diffusion models},
  author={Pandey, Kushagra and Pathak, Jaideep and Xu, Yilun and Mandt, Stephan and Pritchard, Michael and Vahdat, Arash and Mardani, Morteza},
  journal={arXiv preprint arXiv:2410.14171},
  year={2024}
}

@inproceedings{lin2024evaluating,
  title={Evaluating text-to-visual generation with image-to-text generation},
  author={Lin, Zhiqiu and Pathak, Deepak and Li, Baiqi and Li, Jiayao and Xia, Xide and Neubig, Graham and Zhang, Pengchuan and Ramanan, Deva},
  booktitle={European Conference on Computer Vision},
  pages={366--384},
  year={2024},
  organization={Springer}
}

@inproceedings{radford2021learning,
  title={Learning transferable visual models from natural language supervision},
  author={Radford, Alec and Kim, Jong Wook and Hallacy, Chris and Ramesh, Aditya and Goh, Gabriel and Agarwal, Sandhini and Sastry, Girish and Askell, Amanda and Mishkin, Pamela and Clark, Jack and others},
  booktitle={International conference on machine learning},
  pages={8748--8763},
  year={2021},
  organization={PmLR}
}

@article{friedman2022vendi,
  title={The vendi score: A diversity evaluation metric for machine learning},
  author={Friedman, Dan and Dieng, Adji Bousso},
  journal={arXiv preprint arXiv:2210.02410},
  year={2022}
}

@article{stancevic2025information,
  title={The information dynamics of generative diffusion},
  author={Stancevic, Dejan and Ambrogioni, Luca},
  journal={arXiv preprint arXiv:2508.19897},
  year={2025}
}

@article{skreta2025feynman,
  title={Feynman-kac correctors in diffusion: Annealing, guidance, and product of experts},
  author={Skreta, Marta and Akhound-Sadegh, Tara and Ohanesian, Viktor and Bondesan, Roberto and Aspuru-Guzik, Al{\'a}n and Doucet, Arnaud and Brekelmans, Rob and Tong, Alexander and Neklyudov, Kirill},
  journal={arXiv preprint arXiv:2503.02819},
  year={2025}
}

@article{vincent2011connection,
  title={A connection between score matching and denoising autoencoders},
  author={Vincent, Pascal},
  journal={Neural computation},
  volume={23},
  number={7},
  pages={1661--1674},
  year={2011},
  publisher={MIT Press}
}

@inproceedings{corso2024particle,
  title={Particle Guidance: non-{I.I.D.} Diverse Sampling with Diffusion Models},
  author={Corso, Gabriele and Xu, Yilun and De Bortoli, Valentin and Barzilay, Regina and Jaakkola, Tommi},
  booktitle={The Twelfth International Conference on Learning Representations (ICLR)},
  year={2024}
}

@inproceedings{sehwag2022generating,
  title={Generating High Fidelity Data from Low-density Regions using Diffusion Models},
  author={Sehwag, Vikash and Hazirbas, Caner and Gordo, Albert and Ozgenel, Firat and Canton, Cristian},
  booktitle={Proceedings of the IEEE/CVF Conference on Computer Vision and Pattern Recognition (CVPR)},
  year={2022}
}

@article{ho2022classifier,
  title={Classifier-free diffusion guidance},
  author={Ho, Jonathan and Salimans, Tim},
  journal={arXiv preprint arXiv:2207.12598},
  year={2022}
}

@article{heusel2017gans,
  title={Gans trained by a two time-scale update rule converge to a local nash equilibrium},
  author={Heusel, Martin and Ramsauer, Hubert and Unterthiner, Thomas and Nessler, Bernhard and Hochreiter, Sepp},
  journal={Advances in neural information processing systems},
  volume={30},
  year={2017}
}

@inproceedings{pizzi2022self,
  title={A self-supervised descriptor for image copy detection},
  author={Pizzi, Ed and Roy, Sreya Dutta and Ravindra, Sugosh Nagavara and Goyal, Priya and Douze, Matthijs},
  booktitle={Proceedings of the IEEE/CVF Conference on Computer Vision and Pattern Recognition},
  pages={14532--14542},
  year={2022}
}

@article{huang2023t2i,
  title={T2i-compbench: A comprehensive benchmark for open-world compositional text-to-image generation},
  author={Huang, Kaiyi and Sun, Kaiyue and Xie, Enze and Li, Zhenguo and Liu, Xihui},
  journal={Advances in Neural Information Processing Systems},
  volume={36},
  pages={78723--78747},
  year={2023}
}

@inproceedings{peebles2023scalable,
  title={Scalable diffusion models with transformers},
  author={Peebles, William and Xie, Saining},
  booktitle={Proceedings of the IEEE/CVF international conference on computer vision},
  pages={4195--4205},
  year={2023}
}

@inproceedings{rombach2022high,
  title={High-resolution image synthesis with latent diffusion models},
  author={Rombach, Robin and Blattmann, Andreas and Lorenz, Dominik and Esser, Patrick and Ommer, Bj{\"o}rn},
  booktitle={Proceedings of the IEEE/CVF conference on computer vision and pattern recognition},
  pages={10684--10695},
  year={2022}
}

@article{podell2023sdxl,
  title={Sdxl: Improving latent diffusion models for high-resolution image synthesis},
  author={Podell, Dustin and English, Zion and Lacey, Kyle and Blattmann, Andreas and Dockhorn, Tim and M{\"u}ller, Jonas and Penna, Joe and Rombach, Robin},
  journal={arXiv preprint arXiv:2307.01952},
  year={2023}
}

@article{tevet2022human,
  title={Human motion diffusion model},
  author={Tevet, Guy and Raab, Sigal and Gordon, Brian and Shafir, Yonatan and Cohen-Or, Daniel and Bermano, Amit H},
  journal={arXiv preprint arXiv:2209.14916},
  year={2022}
}

@inproceedings{shih2023long,
  title={Long horizon temperature scaling},
  author={Shih, Andy and Sadigh, Dorsa and Ermon, Stefano},
  booktitle={International conference on machine learning},
  pages={31422--31434},
  year={2023},
  organization={PMLR}
}

@inproceedings{esser2024scaling,
  title={Scaling rectified flow transformers for high-resolution image synthesis},
  author={Esser, Patrick and Kulal, Sumith and Blattmann, Andreas and Entezari, Rahim and M{\"u}ller, Jonas and Saini, Harry and Levi, Yam and Lorenz, Dominik and Sauer, Axel and Boesel, Frederic and others},
  booktitle={Forty-first international conference on machine learning},
  year={2024}
}

@article{song2020score,
  title={Score-based generative modeling through stochastic differential equations},
  author={Song, Yang and Sohl-Dickstein, Jascha and Kingma, Diederik P and Kumar, Abhishek and Ermon, Stefano and Poole, Ben},
  journal={arXiv preprint arXiv:2011.13456},
  year={2020}
}

@article{ho2020denoising,
  title={Denoising diffusion probabilistic models},
  author={Ho, Jonathan and Jain, Ajay and Abbeel, Pieter},
  journal={Advances in neural information processing systems},
  volume={33},
  pages={6840--6851},
  year={2020}
}

@article{lipman2022flow,
  title={Flow matching for generative modeling},
  author={Lipman, Yaron and Chen, Ricky TQ and Ben-Hamu, Heli and Nickel, Maximilian and Le, Matt},
  journal={arXiv preprint arXiv:2210.02747},
  year={2022}
}

@article{liu2022flow,
  title={Flow straight and fast: Learning to generate and transfer data with rectified flow},
  author={Liu, Xingchao and Gong, Chengyue and Liu, Qiang},
  journal={arXiv preprint arXiv:2209.03003},
  year={2022}
}

@inproceedings{sohl2015deep,
  title={Deep unsupervised learning using nonequilibrium thermodynamics},
  author={Sohl-Dickstein, Jascha and Weiss, Eric and Maheswaranathan, Niru and Ganguli, Surya},
  booktitle={International conference on machine learning},
  pages={2256--2265},
  year={2015},
  organization={pmlr}
}

@article{goodfellow2014gan,
  title={Generative adversarial nets},
  author={Goodfellow, Ian J and Pouget-Abadie, Jean and Mirza, Mehdi and Xu, Bing and Warde-Farley, David and Ozair, Sherjil and Courville, Aaron and Bengio, Yoshua},
  journal={Advances in neural information processing systems},
  volume={27},
  year={2014}
}

@article{song2020denoising,
  title={Denoising diffusion implicit models},
  author={Song, Jiaming and Meng, Chenlin and Ermon, Stefano},
  journal={arXiv preprint arXiv:2010.02502},
  year={2020}
}

@article{dhariwal2021diffusion,
  title={Diffusion models beat gans on image synthesis},
  author={Dhariwal, Prafulla and Nichol, Alexander},
  journal={Advances in neural information processing systems},
  volume={34},
  pages={8780--8794},
  year={2021}
}

@inproceedings{bansal2023universal,
  title={Universal guidance for diffusion models},
  author={Bansal, Arpit and Chu, Hong-Min and Schwarzschild, Avi and Sengupta, Soumyadip and Goldblum, Micah and Geiping, Jonas and Goldstein, Tom},
  booktitle={Proceedings of the IEEE/CVF conference on computer vision and pattern recognition},
  pages={843--852},
  year={2023}
}

@article{hinton2015distilling,
  title={Distilling the knowledge in a neural network},
  author={Hinton, Geoffrey and Vinyals, Oriol and Dean, Jeff},
  journal={arXiv preprint arXiv:1503.02531},
  year={2015}
}

@article{holtzman2019curious,
  title={The curious case of neural text degeneration},
  author={Holtzman, Ari and Buys, Jan and Du, Li and Forbes, Maxwell and Choi, Yejin},
  journal={arXiv preprint arXiv:1904.09751},
  year={2019}
}

@article{kirkpatrick1983optimization,
  title={Optimization by simulated annealing},
  author={Kirkpatrick, Scott and Gelatt Jr, C Daniel and Vecchi, Mario P},
  journal={science},
  volume={220},
  number={4598},
  pages={671--680},
  year={1983},
  publisher={American association for the advancement of science}
}

@article{nijkamp2023progen2,
  title={Progen2: exploring the boundaries of protein language models},
  author={Nijkamp, Erik and Ruffolo, Jeffrey A and Weinstein, Eli N and Naik, Nikhil and Madani, Ali},
  journal={Cell systems},
  volume={14},
  number={11},
  pages={968--978},
  year={2023},
  publisher={Elsevier}
}

@inproceedings{jayasumana2024rethinking,
  title={Rethinking fid: Towards a better evaluation metric for image generation},
  author={Jayasumana, Sadeep and Ramalingam, Srikumar and Veit, Andreas and Glasner, Daniel and Chakrabarti, Ayan and Kumar, Sanjiv},
  booktitle={Proceedings of the IEEE/CVF conference on computer vision and pattern recognition},
  pages={9307--9315},
  year={2024}
}

@inproceedings{lin2014microsoft,
  title={Microsoft coco: Common objects in context},
  author={Lin, Tsung-Yi and Maire, Michael and Belongie, Serge and Hays, James and Perona, Pietro and Ramanan, Deva and Doll{\'a}r, Piotr and Zitnick, C Lawrence},
  booktitle={European conference on computer vision},
  pages={740--755},
  year={2014},
  organization={Springer}
}

@inproceedings{deng2009imagenet,
  title={Imagenet: A large-scale hierarchical image database},
  author={Deng, Jia and Dong, Wei and Socher, Richard and Li, Li-Jia and Li, Kai and Fei-Fei, Li},
  booktitle={2009 IEEE conference on computer vision and pattern recognition},
  pages={248--255},
  year={2009},
  organization={Ieee}
}
